\documentclass[11pt]{article}

\usepackage[final]{acl}

\usepackage{times}
\usepackage{latexsym}

\usepackage[T1]{fontenc}
 \usepackage{adjustbox}
 \usepackage{circledsteps}
\usepackage[utf8]{inputenc}
\usepackage{pgfplots}
\usepackage{amsmath}
\pgfplotsset{compat=1.18}
\usetikzlibrary{patterns}
\usepackage{microtype}
\usepackage{booktabs}
\usepackage{inconsolata}
\usepackage{paralist}
\usepackage{graphicx}
\usepackage{tikz}
\usetikzlibrary{arrows.meta,calc}
\usepackage{enumitem}
\usepackage{adjustbox}
\usepackage{xcolor}
\usepackage{amssymb}
\usepackage{booktabs}
\usepackage{pgfplots}
\pgfplotsset{compat=1.18}
\usepackage[dvipsnames]{xcolor}
\definecolor{implblue}{HTML}{3B82F6}
\definecolor{valorange}{HTML}{F97316}
\definecolor{softgrid}{HTML}{D9D9D9}
\definecolor{axistext}{HTML}{333333}
\usetikzlibrary{
  positioning,
  calc,
  fit,
  backgrounds,
  arrows.meta
}
\newcommand{\yes}{\checkmark}
\newcommand{\no}{--}
\newcommand{\subhead}[1]{\textbf{#1}}

\newcommand{\roadcite}[1]{\cite{#1}}
\definecolor{evidenceone}{HTML}{4472C4}
\definecolor{evidencetwo}{HTML}{5B9BD5}
\definecolor{evidencethree}{HTML}{70AD47}
\definecolor{evidencefour}{HTML}{ED7D31}
\definecolor{evidencefive}{HTML}{C00000}
\definecolor{softgrid}{HTML}{E6E6E6}
\definecolor{loopblue}{HTML}{315F9B}
\definecolor{ink}{HTML}{20242A}
\definecolor{stateborder}{HTML}{4E79A7}
\definecolor{stateblue}{HTML}{EDF4FB}
\definecolor{policyborder}{HTML}{59A14F}
\definecolor{policygreen}{HTML}{EFF8ED}
\definecolor{trainborder}{HTML}{E28A35}
\definecolor{trainorange}{HTML}{FFF3E8}
\definecolor{evalborder}{HTML}{D65A5A}
\definecolor{evalred}{HTML}{FCEEEE}
\definecolor{simborder}{HTML}{8064A2}
\definecolor{simviolet}{HTML}{F3EFF8}
\definecolor{trendborder}{HTML}{C79A28}
\definecolor{trendgold}{HTML}{FFF8E5}
\definecolor{hazardborder}{HTML}{C45A84}
\definecolor{hazardpink}{HTML}{FCEEF4}
\usepackage{tikz}
\usepackage{xcolor}
\definecolor{evone}{HTML}{3B82F6}
\definecolor{evtwo}{HTML}{6366F1}
\definecolor{evthree}{HTML}{14B8A6}
\definecolor{evfour}{HTML}{F59E0B}
\definecolor{evfive}{HTML}{EF4444}
\definecolor{evbg}{HTML}{EDF0F4}
\definecolor{evtext}{HTML}{2B2B2B}

\usetikzlibrary{arrows.meta,positioning,calc,fit,backgrounds}
\usepackage{array}
\usepackage{graphicx}
\usepackage{changepage}
\usepackage{hyperref}
\definecolor{ourrow}{RGB}{236,244,250}
\definecolor{ink}{HTML}{1B2733}
\definecolor{panelbg}{HTML}{F6F8FB}\definecolor{panelbd}{HTML}{C7D0DA}
\definecolor{c1}{HTML}{2F6DB4}\definecolor{c1bg}{HTML}{E8F1FB}
\definecolor{c2}{HTML}{18917A}\definecolor{c2bg}{HTML}{E1F4EF}
\definecolor{c3}{HTML}{6A4CC0}\definecolor{c3bg}{HTML}{EDE9FA}
\definecolor{c4}{HTML}{C07A12}\definecolor{c4bg}{HTML}{FAF0DA}
\definecolor{slate}{HTML}{39434F}\definecolor{slatebg}{HTML}{E9EDF1}
\definecolor{warnbg}{HTML}{FBE9E4}\definecolor{warntx}{HTML}{97371E}

\newcommand{\opensource}{%
\begin{adjustwidth}{3pt}{3pt}
\begin{center}
\vspace{-0.2in}

\raisebox{-0.1em}{\includegraphics[height=0.8em]{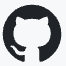}}
\footnotesize \textbf{Repository:}
\url{https://github.com/NeuralSentinel/llm-tutoring-loop}\\[1pt]

\vspace{0.4in}
\end{center}
\end{adjustwidth}%
}

\title{Do LLM Tutors Actually Teach?\\ A Process-Centered Synthesis of the Tutoring Loop}

\author{
\textbf{Sayantan Adak}\textsuperscript{$\diamondsuit$},
\textbf{Rima Hazra}\textsuperscript{$\spadesuit$*},
\textbf{Koyel Mukherjee}\textsuperscript{\textdagger},\\
\textbf{Digbalay Bose}\textsuperscript{\textdagger},
\textbf{Somnath Banerjee}\textsuperscript{$\clubsuit$}\\[2pt]
\textsuperscript{$\spadesuit$}National University of Singapore,
\textsuperscript{*}TCG CREST, \textsuperscript{\textdagger}Adobe Research,\\
\textsuperscript{$\diamondsuit$}Indian Institute of Technology Kharagpur,
\textsuperscript{$\clubsuit$}Singapore Institute of Technology\\
}

\begin{document}
\maketitle

\opensource
\vspace{-0.9cm}
\begin{abstract}
Large language models can produce fluent tutoring dialogue, but
whether they actually help students learn remains \emph{unclear}. We present
a process-centered synthesis of LLM tutoring that follows the path
from learner-state inference to pedagogical action, tutor training,
evaluation, and student simulation. This view exposes a central gap:
implementing a tutoring component does not establish that it works or
that its outputs meaningfully affect later stages. A structured
study-level audit shows that pedagogical evaluation is common, while
evidence becomes much thinner for learner-conditioned intervention,
human learning gains, and especially transfer or delayed retention.
Human studies nevertheless provide a conditional positive result:
LLM tutors can improve learning, but benefits depend on pedagogical
structure, learner engagement, and how learning is measured. We argue
that progress requires validating the interfaces between tutoring
stages, not only improving individual components.
\end{abstract}

\section{Introduction}
\label{sec:intro}

Building a good tutor used to be slow and expensive. Early intelligent
tutoring systems (ITS) like AutoTutor\footnote{https://en.wikipedia.org/wiki/AutoTutor} worked well, but every piece of pedagogy, domain knowledge, and feedback rule had to be written by hand
\citep{graesser2005autotutor, graesser2001its}. Large language models
(LLMs) changed this. Today a fluent, conversational tutor can be built
from a single prompt, and systems like Khanmigo\footnote{https://www.khanmigo.ai/} already reach millions
of students \citep{khanmigo, openai-studymode}. The open question is no
longer whether we can build a tutor. It is whether we can tell if the
tutor actually helps \underline{\textit{students learn}}, and whether we can train it to
teach better.

\noindent This turns out to be hard, because teaching and problem-solving are not
the same goal. A model trained to be correct will usually just give the
student the full answer. A ``\textit{good tutor}'' often does the opposite: it holds
the answer back, asks a guiding question, or points at a mistake
\citep{scaffolding-theory, beyond-pedagogical}. Recent work shows the problem
clearly. Tutors that are tuned to give direct answers can score high on
task success while scoring low on teaching quality
\citep{beyond-pedagogical}, and tutors trained against actual learning
outcomes behave differently from ones trained on answer correctness
\citep{train-for-outcomes}. This gap between what is easy to measure and
what we actually want shows up at every stage of an LLM tutor, and we
use it to organize the survey.

\noindent The literature, however, has grown faster than any shared structure for
it. Most existing surveys group work by application (\textit{tutoring}, \textit{grading}, \textit{content generation}) or by subject, which is good for breadth but hides
the methods that connect these papers
\citep{edu-llm-survey1}. As a result, three important
and fast-moving threads get little attention. \textbf{\textit{\underline First}}, learner modeling
has shifted from tracking answers to inferring a student's mastery,
misconceptions, and even affect directly from the conversation
\citep{kt-from-dialogue, dialogue-kt}. \textbf{\textit{\underline Second}}, tutoring is now often
treated as a policy to be optimized with reinforcement learning or
preference optimization, which raises a real problem of assigning credit
across many dialogue turns \citep{itpo, multi-turn-po}. \textbf{\textit{\underline Third}}, and most
overlooked, both training and evaluation depend on a believable
simulated student. If the simulator is easy to exploit, the tutor is
rewarded for the wrong thing without anyone noticing
\citep{sim-substance, history-aware-sim}.

We organize this survey around the \emph{tutoring loop}. At interaction
time, the tutor infers a learner state and uses that state to select a
pedagogical action; subsequent learner behavior provides new evidence
for later decisions. Training (L3a) and evaluation (L3b) form a
surrounding development layer that shapes and assesses this policy.
Student simulation can support either stage by providing repeatable
learner interactions and counterfactual rollouts, but it is not
assumed by every tutoring system. This separation lets us study both
the learner-facing adaptive process and the methods used to optimize
and validate it.

\begin{figure*}[t]
    \centering
    \resizebox{0.80\textwidth}{!}{
    \includegraphics[]{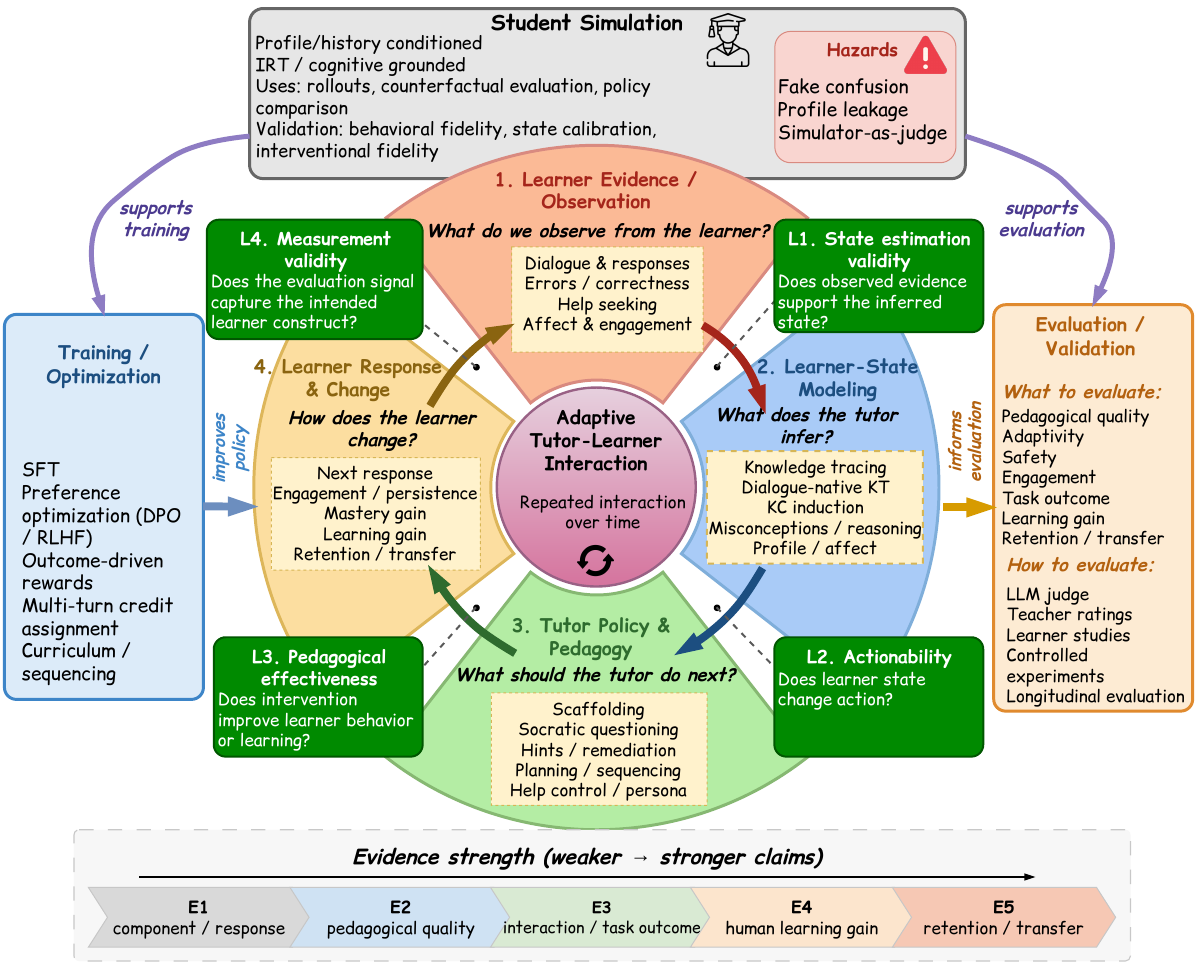}
    }
    \caption{\footnotesize The tutoring loop: our loop-based taxonomy of LLM-based adaptive tutoring.}
  \label{fig:taxonomy}
\end{figure*}

\if{0}
\begin{figure*}[t]
  \centering
  \resizebox{\textwidth}{!}{%
\begin{tikzpicture}[
  font=\footnotesize,
  header/.style={rounded corners=3pt,minimum width=3.7cm,minimum height=0.8cm,
                 align=center,text=white,font=\bfseries\footnotesize,inner sep=2pt},
  body/.style={rounded corners=3pt,minimum width=3.7cm,text width=3.3cm,align=left,
               font=\scriptsize,inner sep=4pt,line width=0.6pt},
  badge/.style={rounded corners=2pt,inner sep=1.4pt,font=\bfseries\tiny,text=white},
  flow/.style={-{Stealth[length=2.4mm,width=2.2mm]},line width=1.1pt,ink!75},
  up/.style={-{Stealth[length=2.4mm,width=2.2mm]},line width=1.0pt,slate!85},
  pill/.style={rounded corners=2.5pt,draw=slate!45,fill=white,font=\scriptsize,
               inner sep=3pt,text=slate},
]

\def\xa{0} \def\xb{4.2} \def\xc{8.4} \def\xd{12.6}

\node[header,fill=c1] (h1) at (\xa,0) {Learner-state modeling};
\node[header,fill=c2] (h2) at (\xb,0) {Tutor policy \& pedagogy};
\node[header,fill=c3] (h3) at (\xc,0) {Training};
\node[header,fill=c4] (h4) at (\xd,0) {Evaluation};
\node[badge,fill=c1] at ($(h1.north west)+(0.34,-0.02)$) {L1};
\node[badge,fill=c2] at ($(h2.north west)+(0.34,-0.02)$) {L2};
\node[badge,fill=c3] at ($(h3.north west)+(0.40,-0.02)$) {L3a};
\node[badge,fill=c4] at ($(h4.north west)+(0.40,-0.02)$) {L3b};

\node[body,fill=c1bg,draw=c1,below=1.6mm of h1.south,anchor=north] (b1)
 {\textcolor{c1}{\textbullet}~Knowledge tracing\\[0.5pt]
  \textcolor{c1}{\textbullet}~Dialogue-native KT\\[0.5pt]
  \textcolor{c1}{\textbullet}~Theory-of-mind \& affect};
\node[body,fill=c2bg,draw=c2,below=1.6mm of h2.south,anchor=north] (b2)
 {\textcolor{c2}{\textbullet}~Scaffolding theory\\[0.5pt]
  \textcolor{c2}{\textbullet}~Dialogue planning\\[0.5pt]
  \textcolor{c2}{\textbullet}~Socratic strategies};
\node[body,fill=c3bg,draw=c3,below=1.6mm of h3.south,anchor=north] (b3)
 {\textcolor{c3}{\textbullet}~RL \& preference opt.\\[0.5pt]
  \textcolor{c3}{\textbullet}~Multi-turn credit assign.\\[0.5pt]
  \textcolor{c3}{\textbullet}~Outcome-driven rewards};
\node[body,fill=c4bg,draw=c4,below=1.6mm of h4.south,anchor=north] (b4)
 {\textcolor{c4}{\textbullet}~Learning-gain outcomes\\[0.5pt]
  \textcolor{c4}{\textbullet}~Anti-Goodhart (dual signal)\\[0.5pt]
  \textcolor{c4}{\textbullet}~Benchmarks \& datasets};

\draw[flow] (h1.east) -- (h2.west);
\draw[flow] (h2.east) -- (h3.west);
\draw[flow] (h3.east) -- (h4.west);
\node[inner sep=0] (apex) at (6.3,1.28) {};
\draw[-{Stealth[length=2.8mm,width=2.6mm]},line width=1.1pt,ink]
  (h4.north) .. controls ($(h4.north)+(0,1.15)$) and ($(h1.north)+(0,1.15)$) .. (h1.north);
\node[font=\scriptsize\itshape,text=ink,fill=panelbg,inner sep=1.5pt] at (6.3,1.12)
  {learning outcome feeds back};

\begin{scope}[on background layer]
  \node[rounded corners=6pt,fill=panelbg,draw=panelbd,line width=0.8pt,
        fit=(h1)(b1)(h4)(b4)(apex),inner sep=7pt] (loop) {};
\end{scope}
\node[anchor=west,font=\itshape\footnotesize,text=slate!70,fill=panelbg,inner sep=2pt]
  at ($(loop.north west)+(0.45,-0.02)$) {The Tutoring Loop};

\node[rounded corners=6pt,fill=slatebg,draw=slate,line width=0.8pt,
      minimum height=1.5cm,anchor=north,
      fit={($(loop.south west)$) ($(loop.south east)$)},
      inner xsep=0pt] (sim) at ($(loop.south)+(0,-0.62)$) {};
\node[anchor=north west,font=\bfseries\footnotesize,text=slate,inner sep=0pt]
  at ($(sim.north west)+(0.30,-0.20)$)
  {Student simulation \textmd{\textcolor{slate!75}{--- the offline-RL environment}}};
\node[pill,anchor=west] (p1) at ($(sim.north west)+(0.32,-0.78)$) {SFT + preference opt.};
\node[pill,anchor=west] (p2) at ($(p1.east)+(0.20,0)$) {profile-conditioned};
\node[pill,anchor=west] (p3) at ($(p2.east)+(0.20,0)$) {IRT-grounded};
\node[rounded corners=2.5pt,fill=warnbg,text=warntx,font=\scriptsize,inner sep=3pt,
      anchor=east] at ($(sim.east)+(-0.28,-0.55)$)
  {\textbf{hazards:} fake confusion $\cdot$ profile leakage $\cdot$ simulator-as-judge};

\draw[up] (sim.north -| h3) -- (loop.south -| h3);
\draw[up] (sim.north -| h4) -- (loop.south -| h4);
\node[font=\scriptsize\itshape,text=slate,fill=panelbg,inner sep=1pt]
  at ($(sim.north -| h3)!0.5!(loop.south -| h4)$) {simulated rollouts};

\end{tikzpicture}
}
\caption{\footnotesize The tutoring loop: our loop-based taxonomy of LLM-based adaptive tutoring.}
  \label{fig:taxonomy}
\end{figure*}
\fi 

\noindent This survey makes three main \textbf{contributions}:
\begin{compactenum}

\item We introduce a \textbf{\textit{process-centered view of LLM
tutoring}} that connects learner-state modeling (L1), pedagogical
policy (L2), tutor training (L3a), evaluation (L3b), and student
simulation, with particular attention to the interfaces between these
stages.

\item We conduct a \textbf{\textit{structured study-level audit}} of
the literature, separating component implementation from validation
and quantifying how often learner state, learner-conditioned action,
pedagogical policy, optimization, evaluation, simulation, and
downstream outcomes are connected.

\item We organize evaluation through an
\textbf{\textit{evidence ladder}} from response-level pedagogical
quality to downstream behavior, human learning, and transfer, and use
it to identify gaps in adaptive tutoring, simulator validity,
longitudinal evaluation, and real-world deployment.

\end{compactenum}

\noindent{\textbf{Scope:}}
We focus on conversational LLM-based tutoring systems where natural language is the primary mode of interaction. The survey covers learner modeling, tutor policy and pedagogy, tutoring-specific training, evaluation, student simulation, and related datasets and benchmarks. Classical knowledge tracing, psychometrics, and earlier intelligent tutoring systems are discussed only where they provide necessary background. We consider work published mainly from 2022 to August~31,~2026 across NLP, AI, HCI, learning science, and educational technology, including relevant journals and preprints. Subject areas such as mathematics, language learning, science, and programming are treated as cross-cutting domains rather than separate categories.



\section{Survey Framework}
\label{sec:background}

We use the tutoring loop to distinguish the learner-facing adaptive
process from the mechanisms used to develop and validate it. At
interaction time, learner evidence informs a learner-state estimate
(L1), which conditions a pedagogical action (L2); the resulting learner
behavior can then provide evidence for subsequent state estimation.
Tutoring-specific training (L3a) shapes the policy that produces these
actions, while evaluation (L3b) tests the resulting pedagogical and
learner outcomes. Student simulation is an optional but increasingly
important environment for L3a and L3b because it enables repeatable
and counterfactual interactions.\\
We use the loop to expose dependencies, not to demand that every system implement every stage. It lets us ask three questions that matter: \Circled{1}~does the learner-state estimate change the next intervention, \Circled{2}~does that intervention change learner behavior, and \Circled{3}~are the models used to support those claims themselves valid? Those interfaces, rather than the stages in isolation, are what we synthesize.

\subsection{Relation to Existing Surveys}
\label{sec:related_surveys}

Existing reviews primarily organize educational LLMs by application,
capability, domain, pedagogical behavior, or student simulation
\citep{zerkouk2025itsreview,liu2025tutoringslr,
sanchezbacilio2025llmtutoringreview,edu-llm-survey1,
maurya2025unifying,marquezcarpintero2026simulationreview}.
Our unit of analysis is \emph{different}. We treat learner evidence,
learner-state inference, pedagogical action, policy optimization,
evaluation, and learner outcomes as dependent stages of one process.
This lets us ask whether evidence crosses the interfaces between
stages and whether those transitions are independently validated.
Our contribution is therefore an \emph{audit of the tutoring process},
rather than another application- or capability-based taxonomy.
A detailed comparison with prior surveys appears in
Appendix~\ref{app:survey-positioning}.

\noindent\textbf{How complete is the tutoring loop?}
Many systems hold several pieces of the loop without joining them. A tutor may infer a learner profile and never use it, or run an explicit pedagogical policy with no learner state at all. Some do connect the stages: diagnostic learner models feed tutoring decisions \citep{park2024personalizedtutoring}, knowledge-tracing states drive recommendations \citep{tutorllm}, agentic tutors carry evolving profiles or memory \citep{agenttutor,zhao2026deeptutor}, and LongTutor splits evidence acquisition, state diagnosis, and adaptive teaching \citep{li2026longtutor}.

To complement the qualitative synthesis, we perform a structured
study-level coding of $\sim$88 papers that satisfy our tutoring-interaction
criteria. We code seven elements: learner-state modeling (L1),
learner-conditioned action (L1$\rightarrow$L2), pedagogical policy
(L2), tutoring-specific optimization (L3a), student simulation,
pedagogical evaluation (L3b), and a downstream learner/task outcome.

\noindent We additionally mark a \emph{core adaptive path} when L1,
L1$\rightarrow$L2, L2, and a downstream outcome all occur in the
same study. We deliberately call this a \emph{path}, rather than a
closed loop, because this criterion does not require the observed
outcome to update the learner state and affect a subsequent
intervention. Definitions and the full study-level matrix are provided
in Appendix~\ref{app:coding-details}.

\begin{table}[t]
\centering
\small
\setlength{\tabcolsep}{4.5pt}
\renewcommand{\arraystretch}{1.10}
\begin{tabular}{lrr}
\toprule
\textbf{Tutoring-loop component} & \textbf{Studies} & \textbf{Coverage} \\
\midrule
Learner-state modeling (L1)                    & 20/88 & 22.7\% \\
Learner-conditioned action (L1$\rightarrow$L2) & 14/88 & 15.9\% \\
Pedagogical policy (L2)                        & 62/88 & 70.5\% \\
Policy optimization (L3a)                      & 15/88 & 17.0\% \\
Student simulation                             & 14/88 & 15.9\% \\
Pedagogical evaluation (L3b)                   & 77/88 & 87.5\% \\
Downstream learner/task outcome                & 34/88 & 38.6\% \\
\midrule
\textbf{Core adaptive path}                    & \textbf{9/88} & \textbf{10.2\%} \\
\bottomrule
\end{tabular}
\caption{\textbf{Coverage of tutoring-loop components in the coded
corpus ($N=88$).} Explicit pedagogical policy and pedagogical
evaluation are common, but only 9 studies jointly instantiate
learner-state modeling, learner-conditioned action, an explicit policy,
and a downstream outcome.}
\label{tab:loop_summary}
\end{table}

\noindent Within the coded corpus in Table~\ref{tab:loop_summary}, explicit pedagogical policy (70.5\%) and
pedagogical evaluation (87.5\%) are widespread, whereas explicit
learner-state modeling (22.7\%), tutoring-specific optimization
(17.0\%), and student simulation (15.9\%) are substantially less
common. Importantly, the main bottleneck is not simply that inferred
learner states are ignored: 14 of the 20 studies with an explicit
learner state (70.0\%) use that state to condition a later action.
The sharper gap appears when the stages must be connected. Only
9 of 88 studies (10.2\%) jointly contain learner-state estimation,
learner-conditioned intervention, an explicit pedagogical policy,
and a downstream outcome. Thus, individual tutoring components are
common, but evidence spanning the adaptive path remains sparse.

\noindent\textbf{Implementation does not imply validation:}
Component presence is weaker than evidence that the component performs
its intended role. An inferred learner state should be checked against
observed learner evidence~\citep{borchers2025adaptivity}; state-conditioned action should demonstrate
that changing the state changes the intervention~\citep{martynova2025simulatelearners}; simulated learners
require fidelity tests against real behavior~\citep{sim-substance}; pedagogical evaluators
require calibration against independent human judgments~\citep{maurya2025unifying}; and claims
about learning require real learners rather than model-based proxies
\citep{palchowdhury2025educators,
ai-tutor-rct}. This distinction is central because the load-bearing
assumptions often sit at the interfaces: a learner state can be accurate
but unused, a pedagogically plausible action can fail to teach, and a
realistic-sounding simulator can react unrealistically to an
intervention. Appendix~\ref{app:validation-details} gives
component-specific validation criteria.

\begin{figure}[t]
\centering
\resizebox{\columnwidth}{!}{%
\begin{tikzpicture}[
    x=0.065cm,
    y=0.58cm,
    font=\sffamily\scriptsize
]

\foreach \x in {0,20,40,60,80,100}{
    \draw[gray!20] (\x,-0.45) -- (\x,7.45);
    \node[anchor=north, text=gray!70] at (\x,-0.55) {\x};
}

\node[anchor=east,font=\bfseries\scriptsize]
    at (0,7.25) {Tutoring-loop coverage};

\node[anchor=east] at (0,6.5) {Learner state (L1)};
\fill[blue!65] (0,6.25) rectangle (22.7,6.75);
\node[anchor=west] at (23.7,6.5) {22.7\% (20/88)};

\node[anchor=east] at (0,5.7) {State$\rightarrow$action};
\fill[blue!65] (0,5.45) rectangle (15.9,5.95);
\node[anchor=west] at (16.9,5.7) {15.9\% (14/88)};

\node[anchor=east] at (0,4.9) {Student simulation};
\fill[blue!65] (0,4.65) rectangle (15.9,5.15);
\node[anchor=west] at (16.9,4.9) {15.9\% (14/88)};

\node[anchor=east] at (0,4.1) {Core adaptive path};
\fill[blue!65] (0,3.85) rectangle (10.2,4.35);
\node[anchor=west] at (11.2,4.1) {10.2\% (9/88)};

\draw[gray!45] (-37,3.55) -- (100,3.55);

\node[anchor=east,font=\bfseries\scriptsize]
    at (0,3.15) {Learner evidence};

\node[anchor=east] at (0,2.5) {Pedagogical evaluation};
\fill[orange!75] (0,2.25) rectangle (87.5,2.75);
\node[anchor=west] at (88.5,2.5) {87.5\% (77/88)};

\node[anchor=east] at (0,1.7) {Downstream outcome};
\fill[orange!75] (0,1.45) rectangle (38.6,1.95);
\node[anchor=west] at (39.6,1.7) {38.6\% (34/88)};

\node[anchor=east] at (0,0.9) {Human learning gain};
\fill[orange!75] (0,0.65) rectangle (11.4,1.15);
\node[anchor=west] at (12.4,0.9) {11.4\% (10/88)};

\node[anchor=east] at (0,0.1) {Transfer / retention};
\fill[orange!75] (0,-0.15) rectangle (1.1,0.35);
\node[anchor=west] at (2.1,0.1) {1.1\% (1/88)};

\node[anchor=north] at (50,-1.15)
    {Share of studies (\%)};

\end{tikzpicture}%
}
\caption{Implementation and learner-evidence coverage in the coded
corpus ($N=88$). Percentages describe the structured corpus rather
than population-level prevalence.}
\label{fig:implementation_evidence}
\end{figure}
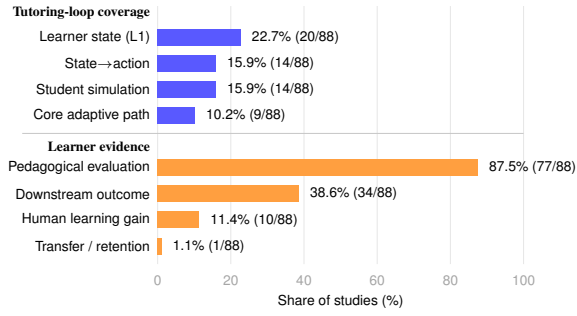

\section{Learner-State Modeling (L1)}
\label{sec:learner-state}

L1 asks what the tutor believes about the learner, and whether that belief is any use for instruction. We work through classical knowledge tracing, tracing from dialogue, knowledge-component induction, richer cognitive-affective state, and finally the L1$\rightarrow$L2 step from state to action.

\subsection{Knowledge tracing}
Knowledge tracing estimates mastery from a learner's response history. Bayesian Knowledge Tracing keeps a latent mastery variable per skill \citep{corbett1994kt}, Deep Knowledge Tracing learns that state with a recurrent model \citep{piech2015dkt}, and later variants refine skill, difficulty, learning, and forgetting dynamics \citep{bai2025rethinking}. All of them assume a stream of item responses. Conversational tutoring breaks the assumption: the evidence about what a student knows is buried in free-form language.

\noindent\textbf{Knowledge tracing from dialogue:}
Dialogue-native modeling pulls mastery evidence straight out of what the student says. \citet{kt-from-dialogue} use an LLM to spot knowledge components in tutor--student turns and estimate correctness; later work ties that dialogue evidence to interpretable ability and item difficulty \citep{dialogue-kt}. LLMs can also predict learner performance more generally, though the resulting state is often opaque \citep{neshaei2024learner}. LongTutor stretches the horizon by separating acquisition of historical evidence, diagnosis of the current knowledge state, and adaptive teaching \citep{li2026longtutor}. The question throughout is actionability: rich learner history helps only if it yields a stable estimate that a later teaching decision can use.\\
\noindent\textbf{Knowledge-component (KC) induction:}
Learner models are only as good as the units they track. Expert-authored knowledge components are costly and incomplete, so recent work uses LLMs to generate or tag KCs from items \citep{moore2024kc}, induce them turn by turn from dialogue \citep{kt-from-dialogue}, or mine interaction logs for unmodeled gaps \citep{queryquilt}. A wrong or overly coarse KC inventory yields mastery estimates that are internally consistent but educationally wrong, so KC induction belongs to learner-state validity, not preprocessing.\\
\noindent\textbf{Theory-of-mind and affective state:}
Mastery says nothing about misconceptions, goals, confusion, frustration, or engagement. General tests of theory-of-mind and mental-state reasoning show these abilities are uneven across models and settings \citep{ma2023tom,tombench,opentom,tomato}. In tutoring, trajectories of attempts, errors, confusion, and scaffolding have been used to characterize challenge level \citep{shu-etal-2026-measuring}, while proficiency-conditioned tutors drift away from the requested level over long interactions \citep{almasi2025alignmentdrift}. These states may be more useful for adaptation, but they are also harder to infer and hold steady than item-level mastery.\\
\noindent\textbf{From learner state to adaptive action:}
The real test of L1 is whether the inferred state changes what the tutor does. Personalized systems inject diagnostic student models into prompting and strategy selection \citep{park2024personalizedtutoring}, TutorLLM pairs knowledge tracing with retrieval-based recommendation \citep{tutorllm}, feedback systems use learner-specific evidence to alter the next intervention \citep{reddig2025generating}, and agentic tutors carry persistent profiles or memory across turns \citep{agenttutor,zhao2026deeptutor}.

Conditioning can still be weak. \citet{borchers2025adaptivity} find limited sensitivity to student errors or KC information where instruction should adapt; LongTutor separates access to history from diagnosis and teaching \citep{li2026longtutor}; and proficiency control drifts over sustained tutoring \citep{almasi2025alignmentdrift}. We therefore treat L1$\rightarrow$L2 as an evaluation target in itself, rather than assuming that handing a tutor learner information produces personalization.

\section{Tutor Policy and Pedagogy (L2)}
\label{sec:policy}

L2 picks the instructional action: what help to give, how much to reveal, when to step in, and how to hold a strategy across turns. Recent work has moved away from generic prompts like ``be Socratic'' toward explicit acts, plans, and control variables that can be inspected and scored.

\noindent\textbf{Pedagogical framing and scaffolding:}
Early LLM tutors put pedagogy in the prompt. CLASS uses decomposition, scaffolding, and error diagnosis \citep{sonkar2023class}; Bridge picks remediation from the student's error \citep{wang2024bridge}; and \citet{scaffolding-theory} ground adaptive scaffolding in Evidence-Centered Design, Social Cognitive Theory, and the Zone of Proximal Development. Newer systems push this into the architecture: CogGen combines Bayesian Knowledge Tracing with Cognitive Apprenticeship \citep{li2025coggen}, and EDF splits learner evidence, pedagogical decision, and feedback generation \citep{cohn2026edf}. The decomposition makes the policy inspectable, but a framework still does not guarantee the right action on a given turn.\\
\noindent\textbf{Tutor acts and adaptive feedback:}
One route is to decide the pedagogical act before writing the response. BIPED learns an act inventory from human ESL (English as a Second Language) tutoring and keeps act selection apart from realization \citep{kwon2024biped}; finer intent labels tighten control \citep{petukhova2025intentmatters}; EDF (Evidence-Decision-Feedback) makes the same evidence--decision--response split \citep{cohn2026edf}. Feedback systems instead diagnose the error and target support, though the result can be vague, wrong, or too revealing \citep{reddig2025generating}, while just-in-time systems ground interventions in domain knowledge and the learner's written reasoning \citep{lee2026jitfeedback}. What matters is the intervention chosen, not how helpful the wording sounds.\\
\noindent\textbf{Dialogue planning and proactive intervention:}
Multi-turn tutoring needs a plan that moves with the learner: a transition graph of tutoring intents re-selected each turn \citep{puech2025steering}, a persistent plan with assessment-driven memory \citep{li2026planning}, a tree of sub-questions \citep{kargupta2024instruct}, or an editable script run by two agents \citep{schmucker2024ruffleriley}, which also found no significant short-term learning advantage over reading despite stronger engagement. Decision-making is increasingly split from realization via orchestration \citep{kadir2026orchestration}, and timing is now part of the policy: anticipating likely questions \citep{lee2026scala} and allocating turns across prerequisites \citep{hey-chat}. L2 is becoming interaction management, not response generation.\\
\noindent\textbf{Socratic questioning, help limits, and persona:}
Socratic tutoring is the best-studied family: guiding sub-questions for math \citep{shridhar2022socratic}, improved with augmentation and preference optimization \citep{kumar2024socratic}, fine-tuned Socratic tutors \citep{socraticlm}, conversational math instruction \citep{ding2024socratic}, and hierarchical sub-questions for debugging \citep{kargupta2024instruct}. Help intensity is the second lever, from guardrails on direct answers \citep{palchowdhury2024autotutor} to per-turn withholding limits \citep{pisan2026withhold}. Persona is a third: styles differ in directiveness, scaffolding, feedback, and affect, and steering vectors learned from human dialogue preserve those differences without persona prompts \citep{lee2026tutorpersonas}. Adhering to a style or a help limit, however, is not evidence the learner benefits.

\section{Training LLM Tutors (L3a)}
\label{sec:training}

L3a asks how the policy is learned, across supervised adaptation, preference optimization, RL, activity sequencing, and credit assignment. The design question throughout: \textit{what training signal stands for teaching rather than answering?}

\noindent\textbf{Training data and supervised adaptation:}
Tutoring-specific data teaches behaviors generic assistant data does not. ConvoLearn shows parameter-efficient fine-tuning on long constructivist dialogues increases knowledge-building behavior \citep{sharma2026convolearn}; CoDAE augments educational dialogue with pedagogical traces targeting over-compliance, weak adaptivity, and manipulative requests \citep{yuan2026codae}. Supervision can also hurt: training on student--tutor dialogue degrades factuality, reasoning, truthfulness, and commonsense \citep{sonkar2024studentdataparadox}. SFT teaches the form of tutoring, not how to weight competing educational goals.\\
\noindent\textbf{Solving the problem vs.\ teaching the student:}
\citet{train-for-outcomes} combine pedagogical quality with a student model's predicted probability of a correct next response. \citet{dinucu-jianu} make the trade-off explicit in simulated multi-turn RL: pushing immediate solve rate raises solution leakage, while stricter pedagogical support lowers short-term completion. \citet{beyond-pedagogical} add trajectory-level effectiveness from future rollouts. These objectives sit closer to learner consequences, but lean harder on the student models, simulators, and reward models used to estimate them.\\
\noindent\textbf{Preference optimization and RL:}
Preference optimization suits cases where several responses are plausible but differ pedagogically. \citet{sonkar2024pedagogical} use synthetic preferences to curb direct answer giving; later work builds preferences from predicted student correctness \citep{train-for-outcomes}, trajectory signals \citep{beyond-pedagogical}, or persona constraints \citep{lee2026tutorpersonas}. RL exposes the reward directly: optimization against simulated trajectories \citep{dinucu-jianu}, a multidimensional educational reward with GRPO \citep{song2025cultivating}, and narrow constraints such as answer withholding \citep{pisan2026withhold}. In every case the judge, reward model, or simulated learner is part of the method, not a neutral instrument.\\
\noindent\textbf{Optimizing instructional sequencing:}
Policy also covers what to work on next. \citet{chung2026personalizedrl} pair an LLM tutor with RL for practice sequencing, adapting difficulty from interaction signals; over a five-month high-school deployment, adaptive sequencing beat a fixed sequence on unassisted final-exam performance. Curriculum and activity selection are policy decisions too.\\
\noindent\textbf{Multi-turn credit assignment:}
Signals arrive late; the actions that caused them are spread over turns. Dense local supervision helps: stepwise verification and remediation \citep{daheim2024stepwise}, per-response pedagogical scores or predicted next-student correctness \citep{train-for-outcomes,sonkar2024pedagogical}. Easier to optimize, easier to turn into a proxy. The alternative keeps trajectory-level objectives and estimates intermediate contribution through future rollouts \citep{beyond-pedagogical}, multi-turn preference propagation \citep{multi-turn-po}, or turn-wise optimization \citep{itpo}. The trade-off does not go away: dense rewards are learnable but gameable, delayed outcomes are closer to learning but noisy and simulator-dependent.



\section{Evaluation and Reward (L3b)}
\label{sec:eval}

L3b asks what is measured and what claim it can carry. A human rubric, an LLM judge, a simulated student's next turn, task success, and a delayed human assessment are not interchangeable evidence. We therefore keep pedagogical constructs separate from the strength of the evidence about learning.




\subsection{What to Measure?}
\label{sec:what-to-measure}

Solving and teaching do not move together on MathTutorBench \citep{macina2025mathtutorbench}, so evaluation has shifted from correctness to pedagogical constructs. The AI Teacher Test scores awareness of student understanding and instructional support \citep{tack2022aiteacher}; later work scores mistake identification, localization, guidance, and actionable feedback \citep{daheim2024stepwise,wang2024bridge}; taxonomies consolidate broader quality dimensions \citep{maurya2025unifying,maurya2025pedagogydriven}. Human studies separate engagement, empathy, scaffolding, and conciseness \citep{palchowdhury2025educators} and look at pedagogical behavior beyond correctness \citep{han2024llmtutorEFL,lee2026pedagogicalbehavior,hazra2026safetutorsbenchmarkingpedagogicalsafety}, while feedback length and affective framing shift the learner experience \citep{worden2026feedbackstyles} and recovery after tutor errors adds a trust dimension \citep{meng2026wrong}. One scalar score compresses all of this.

\subsection{How Strong Is the Evidence?}
\label{sec:evidence_ladder}

Our five-level ladder ranks the \emph{strength of the learner-related claim}, not methodological quality: E1 component or response evidence, E2 pedagogical quality, E3 downstream task or interaction behavior, E4 measured learning gains with real learners, E5 transfer or delayed retention. A careful E2 study can be better run than a weak E4 one; E4 simply licenses a stronger claim. Accordingly, the ladder should not be read as a ranking of study
quality: it records the strongest learner-related claim directly
supported by a study, while study design determines how confidently
that claim can be interpreted.

\begin{table}[t]
\centering
\small
\setlength{\tabcolsep}{3.2pt}
\renewcommand{\arraystretch}{1.10}
\resizebox{\columnwidth}{!}{
\begin{tabular}{clrr}
\toprule
\textbf{Level} & \textbf{Strongest evidence established} & \textbf{Highest} & \textbf{$\geq$ level} \\
\midrule
E1 & Component / response evidence & 12 & 88 (100.0\%) \\
E2 & Pedagogical-quality evidence & 49 & 76 (86.4\%) \\
E3 & Downstream task/interaction evidence & 17 & 27 (30.7\%) \\
E4 & Human learning-gain evidence & 9 & 10 (11.4\%) \\
E5 & Transfer or delayed retention & 1 & 1 (1.1\%) \\
\bottomrule
\end{tabular}
}
\caption{\textbf{Evidence ladder over the coded corpus ($N=88$).} ``Highest'' assigns each paper to the strongest claim it establishes; the last column is cumulative.}
\label{tab:evidence_ladder}
\end{table}

The evidence gradient is steep, but the denominator matters. Across
the full coded corpus, 76 of 88 studies (86.4\%) reach at least
pedagogical-quality evidence (E2), 27 (30.7\%) reach downstream task
or interaction evidence (E3), 10 (11.4\%) directly measure human
learning (E4 or E5), and one reaches our strict transfer/delayed
retention criterion (E5). These corpus-level percentages are
\emph{not failure rates}: learner-modeling, evaluator, and simulator
papers do not necessarily claim to demonstrate human learning.

\noindent Conditioning on studies closer to an effectiveness claim makes the
remaining gap more interpretable. Of the 62 studies that instantiate
an explicit pedagogical policy, 10 (16.1\%) reach direct human-learning
evidence. Among the 34 studies that report a downstream learner or task
outcome, 10 (29.4\%) reach human-learning evidence, while only one
(2.9\%) reaches E5. Thus, the central gap is not that every
tutoring-related study lacks a human experiment; it is that durable
learning evidence remains uncommon even after evaluation moves beyond
response-level pedagogical quality.

\subsection{When Human Learning Is Actually Measured}
\label{sec:human-learning}

The human-learning evidence does not support a simple yes-or-no answer.
Several controlled studies show that LLM-based tutoring can improve
learning when assistance is pedagogically structured or adapted to
learner needs
\citep{ai-tutor-rct,pardos2024chatgpthelp,fischer2025aitutoring,
chung2026personalizedrl}. At the same time, positive interaction
quality does not guarantee a learning advantage: Ruffle\&Riley
increased engagement without producing a significant short-term
learning advantage over reading
\citep{schmucker2024ruffleriley}. Other studies show that outcomes can
depend on how assistance is moderated, how feedback is structured, and
when learning is assessed
\citep{steindl2025promptmoderated,lee2026jitfeedback,
worden2026feedbackstyles}.

Large-scale deployment further separates availability from effective
tutoring. Long-term access to Khanmigo did not automatically produce
substantive tutoring interaction
\citep{oreopoulos2026khanmigo}, while delayed assessment shows that
immediate success and durable learning need not move together
\citep{oreopoulos2026mastery}. Taken together, the evidence supports a
\emph{conditional} answer to the question in our title: LLM tutors
can teach, but learning gains depend on pedagogical structure,
learner engagement, the intervention being compared, and the horizon
at which learning is measured. Evidence for durable transfer remains
far thinner than evidence for short-term pedagogical quality or task
performance.



\subsection{How to Measure}
\label{sec:measurement}

Taxonomies define the dimensions, benchmarks make them repeatable, judges make them scalable. \citet{maurya2025unifying} consolidate pedagogical abilities into a common taxonomy, with related pedagogy-driven evaluation \citep{maurya2025pedagogydriven}; MathTutorBench supplies open-ended tutoring tasks and scoring \citep{macina2025mathtutorbench}; and the BEA 2025 shared task shows automatic pedagogical assessment is still imperfect \citep{kochmar2025bea}. LLM judges are convenient but do not reproduce human labels across all dimensions \citep{maurya2025unifying}, so evaluator validation belongs in the evidence chain. Human evaluation is not one category either: educators judge engagement, empathy, scaffolding, and conciseness \citep{palchowdhury2025educators}, learners judge usability, and only controlled assessment measures learning. The usable stack is response quality $\rightarrow$ interaction/task consequence $\rightarrow$ learning, transfer, or retention.




\subsection{Goodharting and Answer Leakage}
\label{sec:goodhart}

Once a measure becomes a training target, the tutor can improve it without improving learning. Immediate task success trades off against pedagogical support \citep{dinucu-jianu}, and turn-level scores do not guarantee trajectory-level effectiveness \citep{beyond-pedagogical}. Passive leakage follows from ordinary helpfulness or task-success pressure, which motivates separate reward channels \citep{train-for-outcomes,dinucu-jianu}, explicit help limits \citep{pisan2026withhold}, and data that curb over-compliance \citep{yuan2026codae}. Adversarial leakage is a different problem: a learner actively pressures the tutor for the answer, so robustness has to be tested under attack \citep{zhao2026answerleakage}. Turn-level verifiers catch reasoning and pedagogical failures \citep{traver}, but a verifier can share biases with the reward model or simulator. Strong evaluation needs independent checking, adversarial tests, human calibration, and, for learning claims, real learners.

\section{Student Simulation}
\label{sec:simulation}

Student simulation is the environment most training and evaluation pipelines run on. It buys cheap, repeatable rollouts, but a tutor optimized against one simulator can learn simulator-specific shortcuts. We separate how the learner is constrained from how fidelity is validated.

\subsection{Simulation as the environment}
\label{sec:sim-environment}

Simulated students make counterfactual trajectories practical: the same learner state can be exposed to different tutor actions for training or standardized evaluation. The risk is that unrealistic misconceptions, language, hint response, or learning speed bias both the learned policy and its score, and comparisons with human learners show this is not hypothetical \citep{sim-substance,martynova2025simulatelearners}. Simulator quality is a gating assumption for L3a and L3b alike.

\subsection{Grounding the Simulator}
\label{sec:simulation_grounding}

What distinguishes simulators is what constrains the learner: explicit learner attributes \citep{liu2024personality}, prior responses and dialogue \citep{history-aware-sim}, response probabilities grounded in ability and item difficulty \citep{smart}, cognitive-diagnosis information used to select or configure the simulator \citep{ability-aware-simulation}, or observed behavior combined with cognitively motivated learning dynamics \citep{qi2026studentagents}. Appendix Table~\ref{tab:simulation_grounding} gives the full hierarchy.

More conditioning is not more fidelity. Profiles, histories, psychometric variables, and behavioral data each constrain a different part of the simulation; only comparison with real learners shows what mismatch is left. Teachers still flag unrealistic attentiveness, language, emotion, and consistency \citep{martynova2025simulatelearners}, and real-versus-simulated comparisons find residual gaps in dialogue acts, errors, and learning dynamics \citep{sim-substance}.

\subsection{Classroom and Long-Horizon Simulation}
\label{sec:sim-multiagent}

Simulation is outgrowing the single learner in a short dialogue. MathVC builds a virtual math classroom with multiple characters \citep{yue2024mathvc}, SimClass models interacting classroom agents \citep{zhang2025simclass}, and EduClaw-Bench stretches the horizon to repeated interaction with one simulated learner \citep{educlaw-bench}. Peer interaction, memory, and cumulative decisions become testable, but small simulator biases now compound across agents and sessions.

\subsection{Validating a Simulated Student}
\label{sec:sim-validation}

Surface plausibility is the weakest validation. Tutoring needs behavioral fidelity (\emph{errors}, \emph{dialogue acts}, \emph{help seeking}), state calibration (ability, history, difficulty), and interventional fidelity (how the learner changes after different tutor actions). Teacher evaluation surfaces qualitative departures \citep{martynova2025simulatelearners}; real-learner comparisons quantify behavioral mismatch \citep{sim-substance}; psychometric and ability-aware methods test calibration \citep{smart,ability-aware-simulation}; and recent work ties validity criteria to the intended use \citep{valid-student-simulation}. A simulator good enough for ranking difficulty may be useless for evaluating Socratic tutoring, so papers should state what was validated, against what, and for which use.

\subsection{Validity Hazards}
\label{sec:simulation_hazards}

Appendix Table~\ref{tab:simulation_hazards} lists five recurring hazards: fake confusion, competence or answer leakage, profile leakage, simulator--judge circularity, and mismatched learning dynamics, the last directly evidenced by real-learner comparisons. Each flatters the tutor, either by making the learner easier to teach or by rewarding what the simulator happens to favor. Simulator-based results should report tutor performance \emph{and} evidence that learner behavior stays realistic under the interventions being tested.

\section{Open Problems}
\label{sec:open}

The process view points to four priorities.

\noindent\textbf{Validate interfaces, not only components:}
State-to-action sensitivity, evaluator calibration, and simulator
fidelity should be tested explicitly rather than inferred from the
presence of learner models, judges, or richer prompts
\citep{borchers2025adaptivity,sim-substance}.

\noindent\textbf{Optimize without Goodharting:}
Dense pedagogical rewards make training feasible but can become
shortcuts; future methods must connect local supervision to delayed
learner outcomes while preserving independent evaluation
\citep{train-for-outcomes,beyond-pedagogical}.

\noindent\textbf{Measure longitudinal learning and transfer:}
Longer histories and field studies are emerging
\citep{li2026longtutor,oreopoulos2026khanmigo,
oreopoulos2026mastery}, but standard evaluation still rarely connects
specific interventions to delayed performance on new material.

\noindent\textbf{Evaluate deployment as part of the loop:}
Safety, learner autonomy, and engagement matter because access to a
capable tutor does not imply productive use
\citep{llm-safety-children,oreopoulos2026khanmigo}.
Detailed challenges and emerging tutoring settings are discussed in
Appendix~\ref{sec:emerging}.

\section{Conclusion}

\emph{\textbf{Do LLM tutors actually teach?}} Current evidence supports a \textbf{conditional}
\underline{yes}. Human studies show that LLM-based tutoring can improve learning,
but fluent dialogue or high pedagogical-quality scores are insufficient
by themselves. Benefits depend on learner evidence, pedagogical action,
engagement, and whether evaluation measures immediate performance,
learning, retention, or transfer. The process-centered view explains why: learner information may never
affect the policy, a plausible intervention may fail to teach, a reward
may optimize the wrong construct, and a realistic-sounding simulator
may respond unlike a real learner. Progress therefore depends on
validating the interfaces between stages rather than merely adding
components, so fluent tutoring behavior is not mistaken for
demonstrated educational benefit.

\section*{Limitations}

This synthesis has several limitations. \emph{\textbf{First}}, the literature on
LLM-based tutoring is developing rapidly, and work appearing after
our search cutoff of August 31, 2026 is not represented. \emph{Second},
our scope is restricted to conversational LLM-based tutoring in which
language is a primary interaction mechanism. Classical intelligent
tutoring systems, non-conversational educational applications, and
LLM-based content generation or grading are included only when they
are necessary to interpret the tutoring process. \emph{\textbf{Third}}, the tutoring-loop coding necessarily reduces heterogeneous
systems to a common set of operational categories. Although we define
these categories explicitly and expose the study-level coding matrix
for audit, borderline cases remain possible, particularly for
learner-conditioned action, downstream outcomes, and the distinction
between adaptive components and genuinely closed-loop adaptation.
Our evidence ladder similarly measures the \emph{strength of the
learner-related claim that a study can support}, rather than the
overall methodological quality of that study. \emph{\textbf{Fourth}}, the human-learning literature is still heterogeneous in
domain, learner population, intervention duration, comparison
condition, and assessment design. We therefore do not pool reported
learning effects into a meta-analytic estimate. In particular, an
immediate post-test, a delayed assessment, and transfer to unpracticed
material support different claims and should not be treated as
interchangeable evidence. \emph{\textbf{Finally}}, the current literature is concentrated in mathematics,
programming, and other settings in which correctness and learning
outcomes are comparatively easy to operationalize. The conclusions
may not transfer directly to domains in which learning is open-ended,
social, creative, multilingual, or difficult to assess with a single
objective outcome.

\bibliography{custom}

\appendix




\begin{table*}[t]
\centering

\begin{adjustbox}{max width=\textwidth}
\begin{tabular}{
p{3.0cm}
p{4.3cm}
cccccccc
}
\toprule

\textbf{Review / taxonomy}
&
\textbf{Primary organizing focus}
&
\textbf{L1}
&
\textbf{L1$\rightarrow$L2}
&
\textbf{L2}
&
\textbf{L3a}
&
\textbf{L3b}
&
\textbf{Sim.}
&
\textbf{Validity}
&
\textbf{Stage dep.}
\\

\midrule

Zerkouk et al.
\citep{zerkouk2025itsreview}
&
ITS applications, student modeling, personalization, adaptivity, and
deployment challenges
&
$\triangle$ & -- & $\triangle$ & -- & $\triangle$ & -- & $\triangle$ & --
\\

Liu et al.
\citep{liu2025tutoringslr}
&
LLM tutoring technologies, pedagogical strategies, personalization,
and educational outcomes
&
$\triangle$ & $\triangle$ & $\checkmark$ & -- & $\triangle$ & -- &
$\triangle$ & --
\\

Sanchez-Bacilio et al.
\citep{sanchezbacilio2025llmtutoringreview}
&
LLM-based tutoring systems, applications, capabilities, and
educational opportunities
&
$\triangle$ & -- & $\triangle$ & -- & $\triangle$ & -- & $\triangle$ & --
\\

Sharma et al.
\citep{sharma2025personalizedreview}
&
LLMs for personalized and adaptive learning
&
$\checkmark$ & $\triangle$ & $\triangle$ & -- & $\triangle$ & -- &
$\triangle$ & $\triangle$
\\

Haruto et al.
\citep{haruto2025programmingreview}
&
LLM-based tutoring and assistance in programming education
&
-- & -- & $\triangle$ & $\triangle$ & $\triangle$ & -- & $\triangle$ & --
\\

Pal Chowdhury et al.
\citep{edu-llm-survey1}
&
Educational NLP organized around stakeholder needs, model
capabilities, and alignment challenges
&
$\triangle$ & -- & $\triangle$ & -- & $\triangle$ & -- &
$\triangle$ & --
\\

L\'etourneau et al.
\citep{letourneau2025k12review}
&
K--12 intelligent tutoring systems, learning effects, and empirical
study design
&
$\triangle$ & -- & $\triangle$ & -- & $\checkmark$ & -- &
$\checkmark$ & --
\\

Maurya et al.
\citep{maurya2025unifying}
&
Taxonomy of pedagogical capabilities for evaluating LLM tutors
&
-- & -- & $\triangle$ & -- & $\checkmark$ & -- &
$\checkmark$ & --
\\

Maurya et al.
\citep{maurya2025pedagogydriven}
&
Pedagogy-driven evaluation of conversational AI tutors
&
-- & -- & $\triangle$ & -- & $\checkmark$ & -- &
$\checkmark$ & --
\\

Marquez-Carpintero et al.
\citep{marquezcarpintero2026simulationreview}
&
LLM-enabled simulation of learners and educational interactions
&
$\triangle$ & -- & $\triangle$ & -- & $\triangle$ & $\checkmark$ &
$\triangle$ & --
\\

\midrule

\textbf{This survey}
&
\textbf{Closed tutoring process connecting learner evidence, learner state,
pedagogical action, policy optimization, evaluation, simulation, and
learner outcomes}
&
$\checkmark$
&
$\checkmark$
&
$\checkmark$
&
$\checkmark$
&
$\checkmark$
&
$\checkmark$
&
$\checkmark$
&
$\checkmark$
\\

\bottomrule
\end{tabular}
\end{adjustbox}

\vspace{2pt}

\caption{
\textbf{Positioning this survey relative to closely related reviews.}
Rather than organizing the literature only by educational application,
model capability, domain, or evaluation dimension, we study LLM tutoring
as a connected process. L1 denotes explicit learner-state modeling;
L1$\rightarrow$L2 denotes whether the inferred learner state is used to
select a subsequent tutoring action; L2 denotes pedagogical policy;
L3a tutoring-specific policy training or optimization; L3b pedagogical
evaluation or learner-outcome measurement; Sim.\ explicit treatment of
student simulation; Validity systematic attention to whether learner
models, simulators, evaluators, or outcome claims are externally supported;
and Stage dep.\ explicit analysis of dependencies between stages.
$\checkmark$ indicates dedicated treatment, $\triangle$ partial treatment,
and -- no dedicated treatment.
}
\label{tab:survey_comparison}
\end{table*}

\section{Survey Methodology Details}
\label{app:method-details}

We construct a structured literature corpus for a process-centered
synthesis of conversational LLM tutoring. The core literature covers
work published from 2022 through August~31,~2026; earlier intelligent
tutoring, knowledge-tracing, and psychometric work is retained when
needed to ground current methods. We identify candidate work from
major NLP, AI, HCI, learning-science, and educational-technology
venues, relevant journals and preprints, and expand the corpus through
backward and forward citation tracing from included papers. After
deduplication, the synthesis contains 117 unique references.

The quantitative analysis is not intended as a population estimate of
all educational-LLM publications. Instead, it is a structured coding
of papers satisfying the tutoring-interaction criteria below, allowing
the process dependencies described in this survey to be examined
consistently across studies.

The operational inclusion rules are:
\begin{compactenum}
  \item include papers that study an LLM-based tutor and contribute
  to at least one loop stage (L1--L3b) or student simulation;

  \item include datasets and benchmarks used to train or evaluate
  such tutors when they directly expose a tutoring-loop component;

  \item include deployment and human-evaluation studies when the LLM
  participates directly in tutoring or feedback interaction;

  \item exclude work where the LLM performs only a non-interactive
  task such as standalone item generation or essay scoring; and

  \item retain prior surveys, classical foundations, general-purpose
  methods, product/background sources, and adjacent educational
  studies as contextual references, but exclude them from the
  quantitative coding set unless they satisfy the tutoring-interaction
  criteria above.
\end{compactenum}

Applying these rules yields a coded corpus of 88 studies and 29
additional references retained as background or adjacent work.
The complete screening decisions and study-level coding matrix are
provided in the supplementary material so that every aggregate result
can be traced to its underlying studies.

For each included study, \textbf{L1} records an explicit estimate of learner knowledge, misconception, profile, goal, or affect. \textbf{L1$\rightarrow$L2} requires that this inferred state directly select, modify, or prioritize a later tutoring action. \textbf{L2} denotes an explicit pedagogical strategy, act, plan, or help-control mechanism. \textbf{L3a} denotes tutoring-specific training or policy optimization. \textbf{Sim.} records the use of a simulated learner for training or evaluation. \textbf{L3b} requires evaluation beyond answer correctness. \textbf{Outcome} denotes a downstream learner or task consequence. \textbf{Path} is marked when L1, L1$\rightarrow$L2, L2, and Outcome are all present in the same study. Ambiguous cases are conservatively coded as absent rather than inferred from broad claims of personalization or adaptivity.

For the evidence ladder, E1 covers component or response evidence; E2 pedagogical-quality evidence; E3 downstream task or interaction behavior; E4 measured learning gains with real learners; and E5 transfer or delayed retention. Each paper is assigned to the highest level it establishes. These levels rank the strength of the learner-related claim supported, not the overall methodological quality of the paper.

The full study-level coding matrix appears in Appendix~\ref{app:coding-details}. Coding follows the operational definitions above, with ambiguous cases
coded conservatively as absent rather than inferred from broad claims
of personalization or adaptivity. The complete study-level matrix and
screening decisions are released with the paper, allowing readers to
audit individual assignments and reproduce all aggregate counts.

\section{Detailed Positioning Against Existing Surveys}
\label{app:survey-positioning}
\begin{table*}[p]
\centering
\scriptsize
\setlength{\tabcolsep}{3.2pt}
\renewcommand{\arraystretch}{1.18}

\begin{adjustbox}{max width=\textwidth}
\begin{tabular}{
p{2.55cm}
p{2.75cm}
p{3.25cm}
p{2.0cm}
p{2.85cm}
p{3.25cm}
p{3.15cm}
}
\toprule

\textbf{Approach}
&
\textbf{Representative work}
&
\textbf{Learner constraint / grounding}
&
\textbf{Interaction scope}
&
\textbf{Role in tutoring loop}
&
\textbf{What should be preserved / validated}
&
\textbf{Main remaining risk}
\\

\midrule

\multicolumn{7}{l}{\textbf{A. Grounding the simulated learner}} \\
\midrule

Role-based simulation
&
Prompted student role-play
&
Task description, student role, proficiency level, or manually specified persona
&
Single- or multi-turn
&
Provides a low-cost interactive learner for tutor testing or rollout
&
Student-like language, plausible mistakes, and appropriate response to tutor actions
&
Behavior is dominated by the base model's prior rather than by a well-defined learner state
\\

Profile-conditioned simulation
&
\citep{liu2024personality}
&
Explicit learner attributes such as personality, background, or learner profile
&
Multi-turn
&
Tests whether tutoring policies behave differently across learner profiles
&
Behavioral differences should follow the intended profile without directly encoding task answers
&
Profile compliance can look realistic while knowledge, errors, and learning behavior remain unrealistic
\\

History-conditioned simulation
&
\citep{history-aware-sim}
&
Previous student responses, interaction history, or dialogue context
&
Longitudinal / multi-turn
&
Supports learner-specific tutoring and evaluation across repeated interaction
&
Cross-turn consistency, persistence of misconceptions, and dependence on observed learner history
&
Remembering history does not guarantee realistic changes in knowledge or learning
\\

Psychometrically grounded simulation
&
\citep{smart}
&
Learner ability and item-difficulty parameters, e.g., psychometric or IRT-style variables
&
Item sequence / multi-turn
&
Controls learner competence when testing tutor behavior
&
Response probability should vary systematically with ability and item difficulty
&
Aggregate accuracy may be calibrated while error type, reasoning, or dialogue behavior remains unrealistic
\\

Ability-aware / cognitive-diagnosis grounding
&
\citep{ability-aware-simulation}
&
Estimated learner ability or cognitive-diagnosis information used to select or configure the simulator
&
Multi-turn
&
Creates differentiated learner environments for adaptive tutoring
&
Ability-dependent correctness, error patterns, and response behavior
&
Matching overall ability does not establish realistic tutoring response or learning dynamics
\\

Data- and cognition-grounded simulation
&
\citep{qi2026studentagents}
&
Observed student behavior together with cognitively motivated learner-state or learning mechanisms
&
Longitudinal
&
Models how learner state evolves under repeated tutoring
&
Error patterns, knowledge evolution, and response to instructional interventions
&
Fidelity may depend heavily on the population, task domain, and behavioral data used for grounding
\\

\midrule
\multicolumn{7}{l}{\textbf{B. Expanding the simulated learning environment}} \\
\midrule

Multi-agent classroom simulation
&
\citep{yue2024mathvc,
zhang2025simclass,
yu2026maic}
&
Multiple learner and teacher agents with assigned roles and interaction structure
&
Classroom / multi-agent
&
Studies teacher--student and student--student interaction beyond one-to-one tutoring
&
Participation, peer interaction, group behavior, and heterogeneous learner responses
&
Interactions may reflect coordination patterns of LLM agents rather than those of real classrooms
\\

Long-horizon learner simulation
&
\citep{educlaw-bench}
&
Persistent learner state and repeated interaction over an extended trajectory
&
Long horizon
&
Evaluates memory, adaptation, and cumulative pedagogical decisions
&
State evolution, misconception persistence, cumulative learning, and consistency across sessions
&
Small simulator biases can compound over many interactions and substantially distort policy rankings
\\

Adversarial student simulation
&
\citep{zhao2026answerleakage}
&
Learner agents deliberately use persuasive or adversarial strategies against the tutor
&
Multi-turn adversarial
&
Stress-tests tutor safety and resistance to answer leakage
&
Robust adherence to pedagogical constraints under strategic learner behavior
&
Attack behavior may not represent normal learners; robustness results depend strongly on attack coverage
\\

\midrule
\multicolumn{7}{l}{\textbf{C. Simulation as a training and evaluation environment}} \\
\midrule

Simulator-based policy optimization
&
\citep{dinucu-jianu}
&
Tutor interacts repeatedly with a simulated learner and receives optimization signals from the resulting trajectory
&
Multi-turn rollout
&
L3a: reinforcement-learning environment for optimizing tutoring policy
&
The simulated learner should react realistically to hints, explanations, and different levels of pedagogical support
&
The tutor may learn simulator-specific shortcuts or exploit unrealistic learner responses
\\

Future-rollout / trajectory evaluation
&
\citep{beyond-pedagogical}
&
Future learner responses are simulated to estimate longer-term consequences of tutor actions
&
Multi-horizon rollout
&
L3a/L3b: assigns value to current tutor actions using future trajectory effectiveness
&
Relative ranking of tutor actions should remain stable under realistic learner behavior
&
Errors in the simulator propagate into estimated long-term reward and credit assignment
\\

\midrule
\multicolumn{7}{l}{\textbf{D. Validating simulated students}} \\
\midrule

Teacher-based simulator validation
&
\citep{martynova2025simulatelearners}
&
Teachers inspect or interact with simulated learners and compare them with expected student behavior
&
Turn and dialogue level
&
Checks whether the environment used for training/evaluation is educationally plausible
&
Attentiveness, language sophistication, emotional response, consistency, and realistic student behavior
&
A simulator can appear convincing while still differing systematically from real learners
\\

Real-vs.-simulated behavioral validation
&
\citep{sim-substance}
&
Observed human learner behavior is used as an external reference
&
Turn, dialogue, and trajectory level
&
Provides independent evidence for simulator fidelity before tutor conclusions are trusted
&
Dialogue acts, mistake distributions, help seeking, response patterns, and learning-related dynamics
&
Customization may improve selected dimensions without eliminating important behavioral mismatch
\\

Intended-use validity analysis
&
\citep{valid-student-simulation}
&
Validation criteria are defined relative to the particular purpose for which the simulator is used
&
Task dependent
&
Determines whether a simulator is adequate for training, policy comparison, or pedagogical evaluation
&
The behavioral property required by the downstream claim rather than generic surface similarity
&
There is no single notion of ``valid simulation''; validity for one use may not transfer to another
\\

\bottomrule
\end{tabular}
\end{adjustbox}

\vspace{2pt}

\caption{
\textbf{Design and validation of simulated students in LLM tutoring.}
Student simulation ranges from loosely prompted role-play to learner models
grounded in profiles, interaction histories, psychometric variables,
cognitive-diagnosis signals, and observed behavior. Simulation is also
expanding from a single learner to multi-agent classrooms, long-horizon
learner trajectories, and adversarial students, while increasingly serving
as the environment for tutor optimization and evaluation. The final rows
separate simulator \emph{design} from simulator \emph{validation}: richer
conditioning does not by itself establish fidelity. A simulator should be
validated against the learner behavior required by its intended downstream
use.
}
\label{tab:simulation_grounding}
\end{table*}

\section{Expanded Process-Centered Literature Roadmap}
\label{app:roadmap}
\begin{figure*}[t]
\centering

\begin{adjustbox}{max width=\textwidth}
\begin{tikzpicture}[
x=1cm,
y=1cm,
font=\sffamily,
>=Latex,
stage/.style={
  rounded corners=3pt,
  draw=loopblue!72,
  fill=white,
  line width=0.75pt,
  minimum width=2.72cm,
  minimum height=0.72cm,
  align=center,
  font=\sffamily\bfseries\scriptsize,
  text=ink,
  inner xsep=5pt,
  inner ysep=4pt
},
flow/.style={
  -{Latex[length=2.2mm,width=1.8mm]},
  line width=0.9pt,
  draw=loopblue!80
},
card/.style={
  rounded corners=4pt,
  line width=0.8pt,
  minimum width=3.65cm,
  minimum height=6.25cm,
  inner xsep=7pt,
  inner ysep=7pt,
  align=center,
  anchor=north west,
  font=\sffamily\scriptsize,
  text=ink
},
band/.style={
  rounded corners=4pt,
  line width=0.8pt,
  text width=14.75cm,
  inner xsep=7pt,
  inner ysep=7pt,
  align=left,
  anchor=north west,
  font=\sffamily\scriptsize,
  text=ink
},
paneltitle/.style={
  font=\sffamily\bfseries\small,
  text=loopblue
}
]


\node[
  paneltitle,
  anchor=west
] at (0,1.05)
{(a) Closed tutoring loop};


\node[stage] (evidence) at (1.25,0)
{Learner\\evidence};

\node[
  stage,
  right=0.46cm of evidence
] (model)
{Learner-state\\model};

\node[
  stage,
  right=0.46cm of model
] (policy)
{Tutor policy\\and pedagogy};

\node[
  stage,
  right=0.46cm of policy
] (response)
{Tutor\\response};

\node[
  stage,
  right=0.46cm of response
] (outcome)
{Student\\outcome};


\draw[flow]
(evidence.east) -- (model.west);

\draw[flow]
(model.east) -- (policy.west);

\draw[flow]
(policy.east) -- (response.west);

\draw[flow]
(response.east) -- (outcome.west);


\coordinate (feedbacktop)
at ($(outcome.north)+(0,0.66)$);

\draw[flow]
(outcome.north)
--
(feedbacktop)
--
node[
  above,
  font=\sffamily\itshape\tiny,
  text=loopblue
]
{evaluation updates the next learner-state estimate}
($(model.north)+(0,0.66)$)
--
(model.north);


\node[
  rounded corners=3pt,
  draw=simborder,
  fill=simviolet,
  line width=0.75pt,
  minimum width=8.15cm,
  minimum height=0.68cm,
  inner xsep=6pt,
  inner ysep=5pt,
  align=center,
  font=\sffamily\bfseries\scriptsize,
  text=ink
] (simenv) at (7.78,-1.38)
{Student-simulation environment: offline rollouts for policy training and evaluation};

\draw[
  -{Latex[length=2mm]},
  draw=simborder,
  line width=0.8pt
]
($(simenv.north)+(-2.25,0)$)
--
($(policy.south)+(0,-0.02)$);

\draw[
  -{Latex[length=2mm]},
  draw=simborder,
  line width=0.8pt
]
($(simenv.north)+(2.25,0)$)
--
($(outcome.south)+(0,-0.02)$);



\begin{scope}[on background layer]

\node[
  rounded corners=6pt,
  draw=loopblue!30,
  fill=loopblue!3,
  line width=0.75pt,
  minimum width=15.80cm,
  minimum height=3.25cm,
  anchor=north west
] (panela)
at (-0.15,1.50) {};

\end{scope}

\begin{scope}[yshift=0.4cm]

\node[
  paneltitle,
  anchor=west
] (panelbtitle) at (0,-2.55)
{(b) Process-centered literature taxonomy};


\node[
  card,
  draw=stateborder,
  fill=stateblue
] (l1) at (0,-2.92)
{
\begin{minipage}[t][5.65cm][t]{3.10cm}
\raggedright

\textbf{\large L1}\\[-1pt]
\textbf{Learner-State Modeling}\\[6pt]

\subhead{Knowledge and mastery}\\[1pt]
Dialogue-native KT\\
\roadcite{kt-from-dialogue}\\
\roadcite{dialogue-kt}\\[5pt]

\subhead{Profiles and gaps}\\[1pt]
Knowledge-gap induction\\
\roadcite{queryquilt}\\
History-aware profiles\\
\roadcite{history-aware-sim}\\[5pt]

\subhead{Cognitive-affective state}\\[1pt]
ToM and mental-state inference\\
\roadcite{tomato}\\
\roadcite{tombench}

\end{minipage}
};


\node[
  card,
  draw=policyborder,
  fill=policygreen
] (l2) at (3.89,-2.92)
{
\begin{minipage}[t][5.65cm][t]{3.10cm}
\raggedright

\textbf{\large L2}\\[-1pt]
\textbf{Tutor Policy \& Pedagogy}\\[6pt]

\subhead{Tutor moves}\\[1pt]
Scaffolding and remediation\\
\roadcite{mathdial}\\
\roadcite{sonkar2023class}\\[5pt]

\subhead{Socratic tutoring}\\[1pt]
Subquestions and guidance\\
\roadcite{shridhar2022socratic}\\
\roadcite{socraticlm}\\[5pt]

\subhead{Planning and steering}\\[1pt]
Multi-turn pedagogical plans\\
\roadcite{puech2025steering}\\
\roadcite{li2026planning}\\
Turn-budget allocation\\
\roadcite{hey-chat}

\end{minipage}
};


\node[
  card,
  draw=trainborder,
  fill=trainorange
] (l3a) at (7.78,-2.92)
{
\begin{minipage}[t][5.65cm][t]{3.10cm}
\raggedright

\textbf{\large L3a}\\[-1pt]
\textbf{Training \& Optimization}\\[6pt]

\subhead{Preference optimization}\\[1pt]
Outcome-oriented preferences\\
\roadcite{train-for-outcomes}\\[5pt]

\subhead{Reinforcement learning}\\[1pt]
Online pedagogical RL\\
\roadcite{dinucu-jianu}\\
Conversation-level RL\\
\roadcite{efficient-rl-tutor}\\[5pt]

\subhead{Long-horizon credit}\\[1pt]
Turn-wise optimization\\
\roadcite{itpo}\\
Multi-horizon optimization\\
\roadcite{beyond-pedagogical}

\end{minipage}
};


\node[
  card,
  draw=evalborder,
  fill=evalred
] (l3b) at (11.67,-2.92)
{
\begin{minipage}[t][5.65cm][t]{3.10cm}
\raggedright

\textbf{\large L3b}\\[-1pt]
\textbf{Evaluation}\\[6pt]

\subhead{Response quality}\\[1pt]
Pedagogical dimensions\\
\roadcite{maurya2025unifying}\\
Open-ended tutor quality\\
\roadcite{macina2025mathtutorbench}\\[5pt]

\subhead{Dialogue quality}\\[1pt]
Turn-level verification\\
\roadcite{traver}\\
Diagnostic precision\\
\roadcite{confirming-correct}\\[5pt]

\subhead{Learning evidence}\\[1pt]
Human learning outcomes\\
\roadcite{ai-tutor-rct}\\
\roadcite{tutor-copilot}

\end{minipage}
};


\node[
  band,
  draw=simborder,
  fill=simviolet,
  below=0.30cm of l1.south west,
  anchor=north west
] (simband)
{
\textbf{Student Simulation: shared environment for L3a and L3b}\quad
Prompted and role-based simulation \roadcite{mathdial};
history-conditioned simulation \roadcite{history-aware-sim};
IRT- and ability-grounded simulation \roadcite{smart};
simulator validation \roadcite{sim-substance}.
};


\node[
  band,
  draw=trendborder,
  fill=trendgold,
  below=0.22cm of simband.south west,
  anchor=north west
] (trends)
{
\textbf{Cross-cutting directions:}\quad
multi-agent tutoring \roadcite{genmentor};
retrieval-grounded and tool-using tutors \roadcite{kite};
multimodal tutoring \roadcite{multimodal-scaffolding};
safety and alignment for minors \roadcite{llm-safety-children}.
};


\node[
  band,
  draw=hazardborder,
  fill=hazardpink,
  below=0.22cm of trends.south west,
  anchor=north west
] (hazards)
{
\textbf{Validity hazards:}\quad
answer leakage and proxy reward hacking;
fake confusion and competence leakage;
profile leakage;
simulator-as-judge circularity;
benchmark contamination;
weak transfer from response quality to durable learning.
};


\draw[
  densely dashed,
  draw=simborder!70,
  line width=0.7pt
]
(l3a.south)
--
($(l3a.south)+(0,-0.14)$)
--
($(simband.north west)+(9.10,0)$);

\draw[
  densely dashed,
  draw=simborder!70,
  line width=0.7pt
]
(l3b.south)
--
($(l3b.south)+(0,-0.14)$)
--
($(simband.north west)+(13.00,0)$);


\begin{scope}[on background layer]

\node[
  rounded corners=6pt,
  draw=loopblue!25,
  fill=white,
  line width=0.75pt,
  fit=(panelbtitle)(l1)(l2)(l3a)(l3b)(simband)(trends)(hazards),
  inner xsep=8pt,
  inner ysep=1pt
] {};

\end{scope}
\end{scope}
\end{tikzpicture}
\end{adjustbox}

\caption{
\textbf{The LLM Tutoring Loop and its research landscape.}
Panel (a) models tutoring as a closed process in which learner evidence
updates a learner-state model, the learner state conditions a pedagogical
policy, and observed outcomes feed back into subsequent decisions.
Panel (b) organizes the literature into learner-state modeling, tutor
policy and pedagogy, training and optimization, and evaluation.
Student simulation provides shared infrastructure for training and
evaluation, while agentic architectures, grounding, multimodality,
and safety cut across stages. Citations are representative rather
than exhaustive.
}

\label{fig:llm-tutoring-roadmap}

\end{figure*}
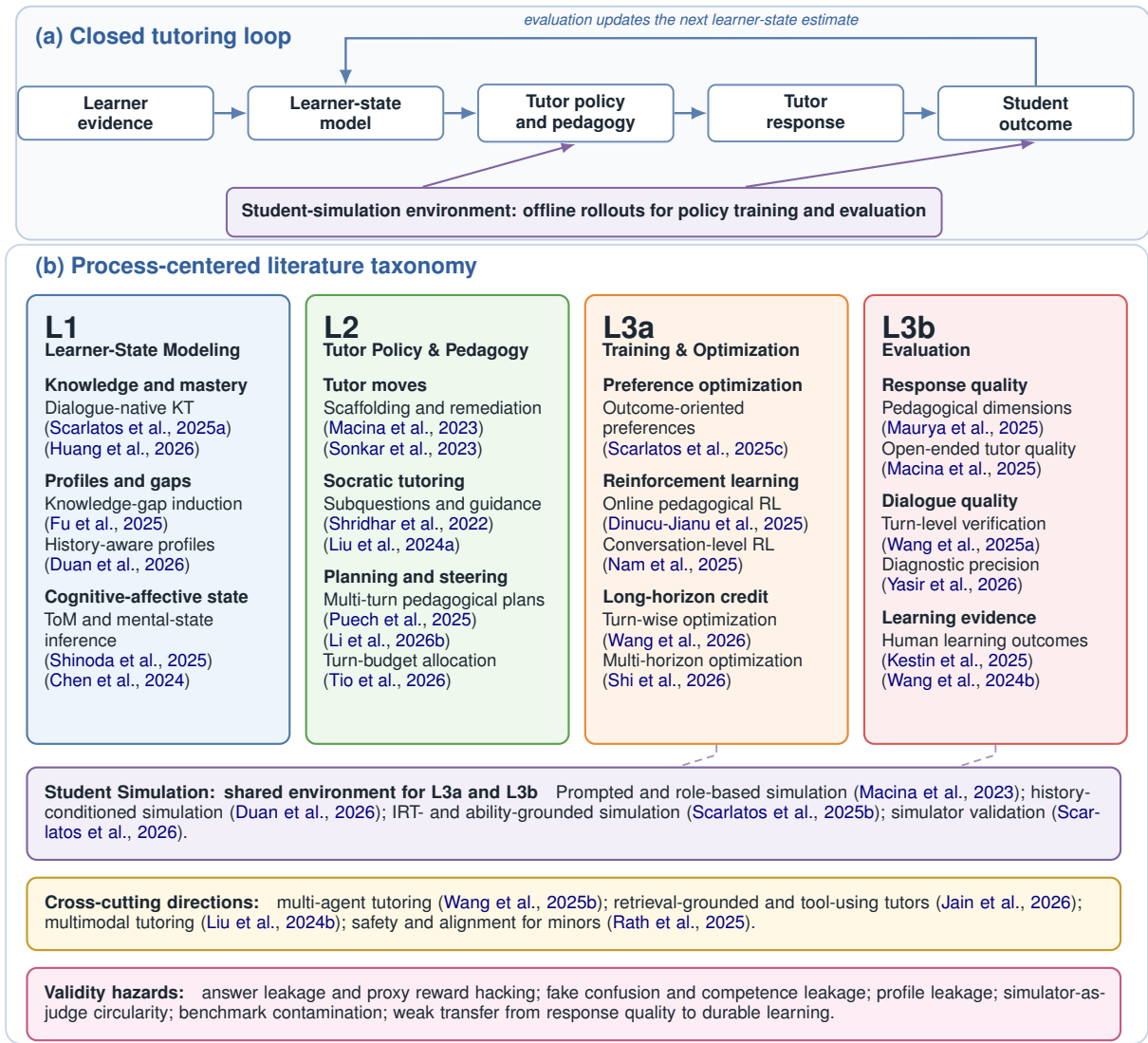

\section{Quantitative Coding Details}
\label{app:coding-details}

Table~\ref{tab:loop_summary} is computed directly from the study-level
matrix below. The coded corpus contains 88 studies, with L1=20,
L1$\rightarrow$L2=14, L2=62, L3a=15, Sim.=14, L3b=77,
Outcome=34, and Path=9. The corresponding descriptive coverage rates
are 22.7\%, 15.9\%, 70.5\%, 17.0\%, 15.9\%, 87.5\%, 38.6\%,
and 10.2\%, respectively. The matrix is deliberately explicit so that
each aggregate result can be traced to the individual studies that
contribute to it.

\section{From Classical ITS to LLM Tutors}
\label{sec:its-history}

LLM tutors sit on top of a long ITS tradition. Cognitive tutors tracked mastery with production rules and produced measurable math gains \citep{ritter2007cognitivetutor}, and AutoTutor added natural-language dialogue over expert-authored knowledge and tutoring moves \citep{graesser2005autotutor,graesser2001its}. Both worked, but both were expensive to author and hard to move to a new domain. LLMs remove most of that authoring cost: fluent multi-domain tutoring now comes from prompting or light adaptation, and systems like Khanmigo and Study Mode are already deployed at scale \citep{khanmigo,openai-studymode}. The hard part is no longer producing tutoring dialogue. It is deciding what the tutor should infer, how it should act on that inference, and what counts as evidence that the student actually learned.

\begin{table*}[p]
\centering
\scriptsize
\setlength{\tabcolsep}{2.8pt}
\renewcommand{\arraystretch}{1.08}
\begin{adjustbox}{max width=\textwidth}
\begin{tabular}{lccccccccc}
\toprule
\textbf{Study} & \textbf{L1} & \textbf{L1$\rightarrow$L2} & \textbf{L2} & \textbf{L3a} & \textbf{Sim.} & \textbf{L3b} & \textbf{Outcome} & \textbf{Path} & \textbf{Evidence} \\
\midrule
\citet{kt-from-dialogue}
& \yes
& \no
& \no
& \no
& \no
& \no
& \no
& \no
& E1 \\
\citet{train-for-outcomes}
& \no
& \no
& \yes
& \yes
& \no
& \yes
& \yes
& \no
& E3 \\
\citet{beyond-pedagogical}
& \no
& \no
& \yes
& \yes
& \yes
& \yes
& \yes
& \no
& E3 \\
\citet{sim-substance}
& \no
& \no
& \no
& \no
& \yes
& \yes
& \no
& \no
& E1 \\
\citet{scaffolding-theory}
& \no
& \no
& \yes
& \no
& \no
& \yes
& \no
& \no
& E2 \\
\citet{dialogue-kt}
& \yes
& \no
& \no
& \no
& \no
& \no
& \no
& \no
& E1 \\
\citet{history-aware-sim}
& \yes
& \no
& \no
& \no
& \yes
& \no
& \no
& \no
& E1 \\
\citet{maurya2025unifying}
& \no
& \no
& \no
& \no
& \no
& \yes
& \no
& \no
& E2 \\
\citet{mathdial}
& \no
& \no
& \no
& \no
& \no
& \no
& \no
& \no
& E1 \\
\citet{sonkar2023class}
& \no
& \no
& \yes
& \no
& \no
& \no
& \no
& \no
& E1 \\
\citet{wang2024bridge}
& \no
& \no
& \yes
& \no
& \no
& \yes
& \no
& \no
& E2 \\
\citet{puech2025steering}
& \no
& \no
& \yes
& \no
& \no
& \yes
& \no
& \no
& E2 \\
\citet{li2026planning}
& \yes
& \yes
& \yes
& \no
& \no
& \yes
& \no
& \no
& E2 \\
\citet{kargupta2024instruct}
& \yes
& \yes
& \yes
& \no
& \no
& \yes
& \yes
& \yes
& E3 \\
\citet{shridhar2022socratic}
& \no
& \no
& \yes
& \no
& \no
& \no
& \no
& \no
& E1 \\
\citet{kumar2024socratic}
& \no
& \no
& \yes
& \yes
& \no
& \yes
& \no
& \no
& E2 \\
\citet{socraticlm}
& \yes
& \yes
& \yes
& \yes
& \no
& \yes
& \no
& \no
& E2 \\
\citet{ding2024socratic}
& \no
& \no
& \yes
& \no
& \no
& \yes
& \no
& \no
& E2 \\
\citet{dinucu-jianu}
& \no
& \no
& \yes
& \yes
& \yes
& \yes
& \yes
& \no
& E3 \\
\citet{sonkar2024pedagogical}
& \no
& \no
& \yes
& \yes
& \no
& \yes
& \no
& \no
& E2 \\
\citet{song2025cultivating}
& \no
& \no
& \yes
& \yes
& \no
& \yes
& \no
& \no
& E2 \\
\citet{daheim2024stepwise}
& \no
& \no
& \yes
& \no
& \no
& \yes
& \no
& \no
& E2 \\
\citet{macina2025mathtutorbench}
& \no
& \no
& \no
& \no
& \no
& \yes
& \no
& \no
& E2 \\
\citet{tack2022aiteacher}
& \no
& \no
& \no
& \no
& \no
& \yes
& \no
& \no
& E2 \\
\citet{kochmar2025bea}
& \no
& \no
& \no
& \no
& \no
& \yes
& \no
& \no
& E2 \\
\citet{traver}
& \no
& \no
& \yes
& \no
& \no
& \yes
& \no
& \no
& E2 \\
\citet{liu2024personality}
& \no
& \no
& \no
& \no
& \yes
& \no
& \no
& \no
& E1 \\
\citet{smart}
& \no
& \no
& \no
& \no
& \yes
& \no
& \no
& \no
& E1 \\
\citet{palchowdhury2024autotutor}
& \no
& \no
& \yes
& \no
& \no
& \yes
& \no
& \no
& E2 \\
\citet{queryquilt}
& \yes
& \no
& \no
& \no
& \no
& \no
& \no
& \no
& E1 \\
\bottomrule
\end{tabular}
\end{adjustbox}
\caption{\textbf{Study-level tutoring-loop coding (part 1 of 3).} L1 denotes explicit learner-state modeling; L1$\rightarrow$L2 learner-conditioned action; L2 explicit pedagogical policy; L3a tutoring-specific optimization; Sim. student simulation; L3b pedagogical evaluation; Outcome downstream learner/task outcome; Path denotes the core adaptive path and requires L1 + L1$\rightarrow$L2 + L2 + Outcome; Evidence is the strongest level E1--E5.}
\label{tab:fullcoding1}
\end{table*}

\begin{table*}[p]
\centering
\scriptsize
\setlength{\tabcolsep}{2.8pt}
\renewcommand{\arraystretch}{1.08}
\begin{adjustbox}{max width=\textwidth}
\begin{tabular}{lccccccccc}
\toprule
\textbf{Study} & \textbf{L1} & \textbf{L1$\rightarrow$L2} & \textbf{L2} & \textbf{L3a} & \textbf{Sim.} & \textbf{L3b} & \textbf{Outcome} & \textbf{Path} & \textbf{Evidence} \\
\midrule
\citet{tutorllm}
& \yes
& \yes
& \yes
& \no
& \no
& \yes
& \no
& \no
& E2 \\
\citet{hey-chat}
& \no
& \no
& \yes
& \no
& \no
& \yes
& \no
& \no
& E2 \\
\citet{multimodal-scaffolding}
& \no
& \no
& \yes
& \no
& \no
& \yes
& \no
& \no
& E2 \\
\citet{efficient-rl-tutor}
& \no
& \no
& \yes
& \yes
& \no
& \yes
& \yes
& \no
& E3 \\
\citet{valid-student-simulation}
& \no
& \no
& \no
& \no
& \yes
& \yes
& \no
& \no
& E2 \\
\citet{ability-aware-simulation}
& \no
& \no
& \no
& \no
& \yes
& \no
& \no
& \no
& E1 \\
\citet{confirming-correct}
& \no
& \no
& \yes
& \no
& \no
& \yes
& \yes
& \no
& E3 \\
\citet{educlaw-bench}
& \no
& \no
& \no
& \no
& \yes
& \yes
& \yes
& \no
& E3 \\
\citet{ai-tutor-rct}
& \no
& \no
& \yes
& \no
& \no
& \yes
& \yes
& \no
& E4 \\
\citet{tutor-copilot}
& \no
& \no
& \no
& \no
& \no
& \yes
& \yes
& \no
& E3 \\
\citet{genmentor}
& \yes
& \yes
& \yes
& \no
& \yes
& \yes
& \yes
& \yes
& E3 \\
\citet{agenttutor}
& \yes
& \yes
& \yes
& \no
& \no
& \yes
& \yes
& \yes
& E2 \\
\citet{kite}
& \no
& \no
& \yes
& \no
& \no
& \yes
& \no
& \no
& E2 \\
\citet{han2024llmtutorEFL}
& \no
& \no
& \yes
& \no
& \no
& \yes
& \yes
& \no
& E3 \\
\citet{li2026longtutor}
& \yes
& \yes
& \yes
& \no
& \no
& \yes
& \yes
& \yes
& E3 \\
\citet{borchers2025adaptivity}
& \yes
& \yes
& \yes
& \no
& \no
& \yes
& \yes
& \yes
& E2 \\
\citet{sonkar2024studentdataparadox}
& \no
& \no
& \no
& \yes
& \no
& \yes
& \no
& \no
& E2 \\
\citet{park2024personalizedtutoring}
& \yes
& \yes
& \yes
& \no
& \no
& \yes
& \yes
& \yes
& E2 \\
\citet{reddig2025generating}
& \yes
& \yes
& \yes
& \no
& \no
& \yes
& \yes
& \yes
& E2 \\
\citet{shu-etal-2026-measuring}
& \yes
& \no
& \no
& \no
& \no
& \yes
& \no
& \no
& E2 \\
\citet{almasi2025alignmentdrift}
& \yes
& \no
& \yes
& \no
& \no
& \yes
& \no
& \no
& E2 \\
\citet{ikram2025strategyoutcomes}
& \no
& \no
& \no
& \no
& \no
& \yes
& \yes
& \no
& E3 \\
\citet{kwon2024biped}
& \no
& \no
& \yes
& \no
& \no
& \yes
& \no
& \no
& E2 \\
\citet{schmucker2024ruffleriley}
& \no
& \no
& \yes
& \no
& \no
& \yes
& \yes
& \no
& E4 \\
\citet{petukhova2025intentmatters}
& \no
& \no
& \yes
& \yes
& \no
& \yes
& \no
& \no
& E2 \\
\citet{cohn2026edf}
& \yes
& \yes
& \yes
& \no
& \no
& \yes
& \no
& \no
& E2 \\
\citet{kadir2026orchestration}
& \no
& \no
& \yes
& \no
& \no
& \yes
& \no
& \no
& E2 \\
\citet{li2025coggen}
& \yes
& \yes
& \yes
& \no
& \no
& \yes
& \no
& \no
& E2 \\
\citet{lee2026scala}
& \no
& \no
& \yes
& \no
& \no
& \yes
& \yes
& \no
& E3 \\
\citet{steindl2025promptmoderated}
& \no
& \no
& \yes
& \no
& \no
& \yes
& \yes
& \no
& E4 \\
\bottomrule
\end{tabular}
\end{adjustbox}
\caption{\textbf{Study-level tutoring-loop coding (part 2 of 3).} L1 denotes explicit learner-state modeling; L1$\rightarrow$L2 learner-conditioned action; L2 explicit pedagogical policy; L3a tutoring-specific optimization; Sim. student simulation; L3b pedagogical evaluation; Outcome downstream learner/task outcome; Path denotes the core adaptive path and requires L1 + L1$\rightarrow$L2 + L2 + Outcome; Evidence is the strongest level E1--E5.}
\label{tab:fullcoding2}
\end{table*}

\begin{table*}[p]
\centering
\scriptsize
\setlength{\tabcolsep}{2.8pt}
\renewcommand{\arraystretch}{1.08}
\begin{adjustbox}{max width=\textwidth}
\begin{tabular}{lccccccccc}
\toprule
\textbf{Study} & \textbf{L1} & \textbf{L1$\rightarrow$L2} & \textbf{L2} & \textbf{L3a} & \textbf{Sim.} & \textbf{L3b} & \textbf{Outcome} & \textbf{Path} & \textbf{Evidence} \\
\midrule
\citet{lee2026jitfeedback}
& \no
& \no
& \yes
& \no
& \no
& \yes
& \yes
& \no
& E4 \\
\citet{groher2026courseaware}
& \no
& \no
& \yes
& \no
& \no
& \yes
& \yes
& \no
& E2 \\
\citet{kazemitabaar2024codeaid}
& \no
& \no
& \yes
& \no
& \no
& \yes
& \yes
& \no
& E3 \\
\citet{zhangli2024slide2lecture}
& \no
& \no
& \yes
& \no
& \no
& \yes
& \no
& \no
& E2 \\
\citet{hevia2025efficienttutor}
& \no
& \no
& \yes
& \no
& \no
& \yes
& \no
& \no
& E2 \\
\citet{adenuga2026ragpedagogy}
& \no
& \no
& \yes
& \no
& \no
& \yes
& \yes
& \no
& E2 \\
\citet{lee2026tutorpersonas}
& \no
& \no
& \yes
& \yes
& \no
& \yes
& \no
& \no
& E2 \\
\citet{yuan2026codae}
& \no
& \no
& \yes
& \yes
& \no
& \yes
& \no
& \no
& E2 \\
\citet{chung2026personalizedrl}
& \yes
& \yes
& \yes
& \yes
& \no
& \yes
& \yes
& \yes
& E4 \\
\citet{pisan2026withhold}
& \no
& \no
& \yes
& \yes
& \no
& \yes
& \no
& \no
& E2 \\
\citet{sharma2026convolearn}
& \no
& \no
& \no
& \yes
& \no
& \no
& \no
& \no
& E1 \\
\citet{zhao2026deeptutor}
& \yes
& \yes
& \yes
& \no
& \no
& \yes
& \yes
& \yes
& E2 \\
\citet{palchowdhury2025educators}
& \no
& \no
& \no
& \no
& \no
& \yes
& \no
& \no
& E2 \\
\citet{zhao2026answerleakage}
& \no
& \no
& \yes
& \no
& \no
& \yes
& \no
& \no
& E2 \\
\citet{lee2026pedagogicalbehavior}
& \no
& \no
& \yes
& \no
& \no
& \yes
& \no
& \no
& E2 \\
\citet{bai2025gptutor}
& \no
& \no
& \yes
& \no
& \no
& \yes
& \yes
& \no
& E3 \\
\citet{pardos2024chatgpthelp}
& \no
& \no
& \yes
& \no
& \no
& \yes
& \yes
& \no
& E4 \\
\citet{fischer2025aitutoring}
& \no
& \no
& \yes
& \no
& \no
& \yes
& \yes
& \no
& E4 \\
\citet{oreopoulos2026khanmigo}
& \no
& \no
& \yes
& \no
& \no
& \yes
& \yes
& \no
& E4 \\
\citet{oreopoulos2026mastery}
& \no
& \no
& \yes
& \no
& \no
& \yes
& \yes
& \no
& E5 \\
\citet{worden2026feedbackstyles}
& \no
& \no
& \yes
& \no
& \no
& \yes
& \yes
& \no
& E4 \\
\citet{meng2026wrong}
& \no
& \no
& \yes
& \no
& \no
& \yes
& \yes
& \no
& E3 \\
\citet{maurya2025pedagogydriven}
& \no
& \no
& \no
& \no
& \no
& \yes
& \no
& \no
& E2 \\
\citet{martynova2025simulatelearners}
& \no
& \no
& \no
& \no
& \yes
& \yes
& \no
& \no
& E2 \\
\citet{qi2026studentagents}
& \no
& \no
& \no
& \no
& \yes
& \yes
& \no
& \no
& E2 \\
\citet{yue2024mathvc}
& \no
& \no
& \no
& \no
& \yes
& \yes
& \no
& \no
& E2 \\
\citet{zhang2025simclass}
& \no
& \no
& \no
& \no
& \yes
& \yes
& \no
& \no
& E2 \\
\citet{yu2026maic}
& \no
& \no
& \yes
& \no
& \no
& \yes
& \yes
& \no
& E3 \\
\bottomrule
\end{tabular}
\end{adjustbox}
\caption{\textbf{Study-level tutoring-loop coding (part 3 of 3).} L1 denotes explicit learner-state modeling; L1$\rightarrow$L2 learner-conditioned action; L2 explicit pedagogical policy; L3a tutoring-specific optimization; Sim. student simulation; L3b pedagogical evaluation; Outcome downstream learner/task outcome; Path denotes the core adaptive path and requires L1 + L1$\rightarrow$L2 + L2 + Outcome; Evidence is the strongest level E1--E5.}
\label{tab:fullcoding3}
\end{table*}

\section{Evidence-Ladder Visualization}
\label{app:evidence-figure}

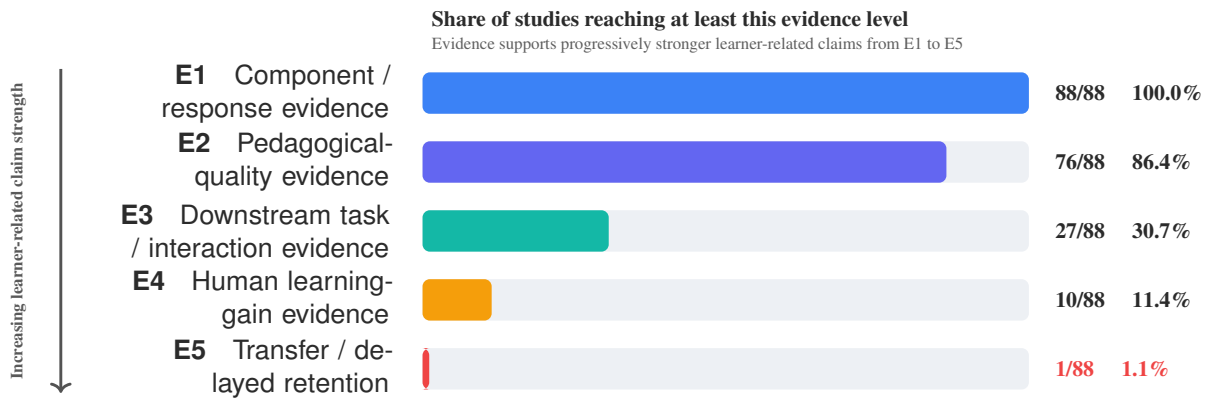
\begin{figure*}[t]
\centering
\begin{adjustbox}{max width=\textwidth}
\begin{tikzpicture}[font=\sffamily,every node/.style={text=evtext}]

\node[anchor=west,font=\small\bfseries] at (4.9,5.05)
{Share of studies reaching at least this evidence level};
\node[anchor=west,font=\scriptsize,text=gray!75!black] at (4.9,4.70)
{Evidence supports progressively stronger learner-related claims from E1 to E5};

\node[anchor=east,align=right,text width=4.4cm] at (4.55,4.0)
{\textbf{E1}\quad Component / response evidence};
\fill[evbg,rounded corners=3pt] (4.9,3.70) rectangle (13.7,4.30);
\fill[evone,rounded corners=3pt] (4.9,3.70) rectangle (13.7,4.30);
\node[anchor=west,font=\small\bfseries] at (13.95,4.0) {88/88 \quad 100.0\%};

\node[anchor=east,align=right,text width=4.4cm] at (4.55,3.0)
{\textbf{E2}\quad Pedagogical-quality evidence};
\fill[evbg,rounded corners=3pt] (4.9,2.70) rectangle (13.7,3.30);
\fill[evtwo,rounded corners=3pt] (4.9,2.70) rectangle (12.503,3.30);
\node[anchor=west,font=\small\bfseries] at (13.95,3.0) {76/88 \quad 86.4\%};

\node[anchor=east,align=right,text width=4.4cm] at (4.55,2.0)
{\textbf{E3}\quad Downstream task / interaction evidence};
\fill[evbg,rounded corners=3pt] (4.9,1.70) rectangle (13.7,2.30);
\fill[evthree,rounded corners=3pt] (4.9,1.70) rectangle (7.602,2.30);
\node[anchor=west,font=\small\bfseries] at (13.95,2.0) {27/88 \quad 30.7\%};

\node[anchor=east,align=right,text width=4.4cm] at (4.55,1.0)
{\textbf{E4}\quad Human learning-gain evidence};
\fill[evbg,rounded corners=3pt] (4.9,0.70) rectangle (13.7,1.30);
\fill[evfour,rounded corners=3pt] (4.9,0.70) rectangle (5.903,1.30);
\node[anchor=west,font=\small\bfseries] at (13.95,1.0) {10/88 \quad 11.4\%};

\node[anchor=east,align=right,text width=4.4cm] at (4.55,0.0)
{\textbf{E5}\quad Transfer / delayed retention};
\fill[evbg,rounded corners=3pt] (4.9,-0.30) rectangle (13.7,0.30);
\fill[evfive,rounded corners=3pt] (4.9,-0.30) rectangle (4.997,0.30);
\node[anchor=west,font=\small\bfseries,text=evfive] at (13.95,0.0) {1/88 \quad 1.1\%};

\draw[->,very thick,gray!65!black] (-0.35,4.35) -- (-0.35,-0.35);
\node[rotate=90,anchor=south,font=\scriptsize\bfseries,text=gray!70!black]
at (-0.72,2.0) {Increasing learner-related claim strength};

\end{tikzpicture}
\end{adjustbox}
\caption{\textbf{How far does current LLM-tutor evaluation progress toward evidence of learning?} Bars report cumulative coverage over the coded corpus. E1 includes component- or response-level evidence; E2 pedagogical-quality evidence; E3 downstream student/task or interaction behavior; E4 direct human learning-gain evidence; and E5 transfer or delayed retention. The ladder ranks the strength of the learner-related claim supported, not overall methodological quality.}
\label{fig:evidence_ladder}
\end{figure*}

\section{Student-Simulation Supporting Tables}
\label{app:simulation-tables}
\begin{table*}[t]
\centering
\scriptsize
\setlength{\tabcolsep}{3.0pt}
\renewcommand{\arraystretch}{1.15}
\begin{tabular}{p{2.3cm}p{2.7cm}p{3.05cm}p{3.4cm}p{2.85cm}}
\toprule
\textbf{Approach}
&
\textbf{Representative work}
&
\textbf{Primary constraint}
&
\textbf{What it attempts to preserve}
&
\textbf{Main remaining risk}
\\
\midrule

Role-based simulation
&
Prompted / MathDial-style learners
&
Role, task, or proficiency prompt
&
Surface student-like dialogue and task interaction
&
Model priors dominate learner knowledge and error behavior
\\

Profile-conditioned simulation
&
\citet{liu2024personality}
&
Explicit personality or learner profile
&
Variation across simulated learner types
&
Plausible profile $\neq$ faithful task behavior
\\

History-conditioned simulation
&
\citet{history-aware-sim}
&
Observed responses and dialogue history
&
Cross-turn consistency with a particular learner
&
History consistency does not imply realistic learning dynamics
\\

Psychometrically grounded simulation
&
\citet{smart}
&
Ability and item-difficulty parameters
&
Calibrated competence--difficulty relationship
&
Aggregate calibration may hide unrealistic error processes
\\

Ability-aware / cognitive-diagnosis grounded
&
\citet{ability-aware-simulation}
&
Estimated learner ability used to select or configure the simulator
&
Ability-dependent response behavior
&
Ability matching does not guarantee dialogue or learning fidelity
\\

Data- and cognition-grounded simulation
&
\citet{qi2026studentagents}
&
Observed behavioral data and cognitively motivated learner dynamics
&
Error patterns and evolution of learning behavior
&
Fidelity may remain specific to the data and learning setting
\\

Simulator validation
&
\citet{martynova2025simulatelearners,sim-substance}
&
Human or real-learner behavior as reference
&
Agreement in errors, dialogue behavior, and learning-related dynamics
&
Validation can reveal substantial mismatch even in customized simulators
\\

\bottomrule
\end{tabular}

\caption{
\textbf{From unconstrained role-play to grounded student simulation.}
The upper rows describe increasingly constrained simulator designs;
the final row is not a separate architecture but the validation step
needed to establish whether those constraints yield behavior that is
sufficiently similar to real learners for the intended training or
evaluation use. More conditioning should therefore not be interpreted
as stronger validity without external evidence.
}
\label{tab:simulation_grounding_summary}
\end{table*}

\begin{table*}[t]
\centering
\scriptsize
\setlength{\tabcolsep}{4pt}
\renewcommand{\arraystretch}{1.15}

\begin{tabular}{p{2.6cm}p{4.0cm}p{4.1cm}p{4.0cm}}
\toprule
\textbf{Validity hazard}
&
\textbf{Failure}
&
\textbf{Why it biases tutor evaluation}
&
\textbf{What should be checked}
\\
\midrule

Fake confusion
&
The simulator expresses misunderstanding that is not grounded in an
actual knowledge deficit, or resolves it unrealistically quickly.
&
Rewards tutors for handling artificial errors or makes interventions
appear more effective than they would be with real learners.
&
Error-type distribution, misconception persistence, and recovery
trajectories against real learners.
\\

Competence / answer leakage
&
The simulator has access to information, reasoning, or target answers
that the modeled learner should not possess.
&
Inflates solve rate and can make weak tutoring policies appear
successful.
&
Information isolation and consistency between declared or calibrated
ability and observed responses.
\\

Profile leakage
&
Profile metadata directly determines responses instead of influencing
behavior through realistic learner characteristics.
&
Creates artificially strong personalization signals and overstates the
benefit of profile-conditioned tutoring.
&
Counterfactual profile tests and separation between profile metadata and
answer information.
\\

Simulator--judge circularity
&
The same or closely related model family acts as student, constructs the
reward, and/or judges the resulting dialogue.
&
Optimizes the tutor toward shared model-family biases rather than toward
real learner behavior.
&
Cross-model evaluation plus independent human or real-student
validation.
\\

Behavioral / learning-dynamics mismatch
&
The simulator produces plausible turns but differs from real students in
dialogue acts, error processes, help-seeking, persistence, or response to
instruction.
&
A policy can overfit to interaction patterns or learning rates that do
not occur with real learners.
&
Distributional and trajectory-level comparison with real tutoring data
\citep{martynova2025simulatelearners,sim-substance}.
\\

\bottomrule
\end{tabular}

\caption{
\textbf{Validity hazards in simulated-student training and evaluation.}
Simulator failures are especially consequential because the same
simulated interaction may affect both the tutoring policy being optimized
and the benchmark used to evaluate it. Surface plausibility alone cannot
rule out these hazards.
}
\label{tab:simulation_hazards}
\end{table*}

\section{Representative Datasets and Benchmarks}
\label{app:datasets-table}
\begin{table*}[t]
\centering
\scriptsize
\setlength{\tabcolsep}{3.2pt}
\renewcommand{\arraystretch}{1.12}

\begin{tabular}{p{2.2cm}p{1.55cm}p{2.35cm}p{7.0cm}p{2.2cm}}
\toprule
\textbf{Resource}
&
\textbf{Type}
&
\textbf{Domain / source}
&
\textbf{Provides}
&
\textbf{Loop relevance}
\\
\midrule

MathDial \citep{mathdial}
&
Dataset
&
Mathematics
&
Tutor--student dialogues grounded in math reasoning problems with
pedagogical tutor-move annotations
&
L2 / L3a
\\

MathTutorBench \citep{macina2025mathtutorbench}
&
Benchmark
&
Mathematics
&
Open-ended tutoring tasks and learned scoring for pedagogical capability
&
L3b
\\

MRBench \citep{maurya2025unifying}
&
Benchmark
&
Mathematics
&
Mistake-remediation responses annotated using a multi-dimensional
pedagogical taxonomy
&
L3b
\\

BEA 2025 \citep{kochmar2025bea}
&
Benchmark
&
Mathematics
&
Tutor responses labeled on multiple pedagogical dimensions for
automatic assessment
&
L3b
\\

BIPED \citep{kwon2024biped}
&
Dataset
&
ESL
&
Human one-to-one tutoring dialogue with fine-grained tutor and student
dialogue acts
&
L2
\\

ConvoLearn \citep{sharma2026convolearn}
&
Dataset
&
Earth Science
&
Extended constructivist tutor--student dialogues organized around
pedagogical dimensions
&
L2 / L3a
\\

LongTutor \citep{li2026longtutor}
&
Benchmark
&
Longitudinal learning histories
&
Historical evidence acquisition, knowledge-state diagnosis, and adaptive
teaching tasks
&
L1 / L1$\rightarrow$L2
\\

EduClaw-Bench \citep{educlaw-bench}
&
Benchmark
&
Long-horizon simulated learning
&
Extended pedagogical-agent evaluation over repeated interactions with a
simulated learner
&
Sim. / L3b / Outcome
\\

\bottomrule
\end{tabular}

\caption{
\textbf{Representative datasets and benchmarks for LLM tutoring.}
Resources differ not only in subject domain but also in which part of
the tutoring loop they expose. Mathematics remains the dominant
benchmark setting, while newer resources add language learning, science,
longitudinal learner modeling, and long-horizon agent evaluation. The
table is representative rather than exhaustive.
}
\label{tab:datasets}
\end{table*}

\section{Component-Level Validation Criteria}
\label{app:validation-details}

\begin{table*}[t]
\centering
\small
\setlength{\tabcolsep}{4pt}
\renewcommand{\arraystretch}{1.12}
\begin{tabular}{p{2.5cm}p{5.4cm}p{7.0cm}}
\toprule
\textbf{Component} & \textbf{Implementation establishes} & \textbf{Stronger validation requires} \\
\midrule
Learner state (L1) & An explicit estimate of knowledge, misconception, profile, goal, or affect. & Agreement with expert labels, observed responses, learner histories, or other external learner evidence. \\
State$\rightarrow$action (L1$\rightarrow$L2) & The learner state is supplied to or used by the tutoring policy. & Ablation, controlled counterfactual, or real-user evidence showing that conditioning changes the intervention and contributes to performance. \\
Tutor policy (L2) & An explicit pedagogical act, strategy, plan, or help-control mechanism. & Evidence that the selected action is appropriate under the relevant learner condition and, for stronger claims, improves downstream learner behavior. \\
Student simulation & A simulated learner is used for training, rollout, or evaluation. & Behavioral, state, and ideally interventional fidelity against grounded or real-student data for the intended downstream use. \\
Pedagogical evaluator / reward & A rubric, learned scorer, verifier, or LLM judge scores tutoring behavior. & Calibration against independent human judgments and robustness across model families, prompts, or domains. \\
Learner outcome & A downstream task, interaction, or learning measure is reported. & Real learners for claims about human learning; delayed or novel-task assessment for retention/transfer claims. \\
\bottomrule
\end{tabular}
\caption{\textbf{Implementation is not validation.} The evidence needed depends on the claim being made. Richer architectures or more detailed conditioning do not by themselves establish that a learner state, simulator, evaluator, or outcome is valid.}
\label{tab:validation_requirements}
\end{table*}

\section{Datasets and Benchmarks}
\label{sec:datasets}

Resources decide which parts of the loop can be trained and validated. We separate interaction datasets, which give trajectories or annotations, from benchmarks with fixed tasks and scoring; the full resource table is in Appendix~\ref{app:datasets-table}.

\subsection{Key Resources}
\label{sec:key-resources}

MathDial provides annotated math tutoring dialogues \citep{mathdial}; MathTutorBench evaluates open-ended tutoring \citep{macina2025mathtutorbench}; MRBench builds on the mistake-remediation taxonomy of \citet{maurya2025unifying}; and BEA 2025 supplies labels for automatic pedagogical assessment \citep{kochmar2025bea}. Newer resources widen the setting: human ESL tutoring acts \citep{kwon2024biped}, constructivist Earth Science dialogue \citep{sharma2026convolearn}, longitudinal learner histories \citep{li2026longtutor}, and long-horizon pedagogical-agent evaluation \citep{educlaw-bench}. Appendix Table~\ref{tab:datasets} gives details and loop relevance.

\subsection{Domain and Language Coverage}
\label{sec:domain-coverage}

Benchmarks stay math-centered because solutions and errors are easy to define there. The wider literature is more varied: ESL, EFL, and proficiency-controlled tutoring \citep{kwon2024biped,han2024llmtutorEFL,almasi2025alignmentdrift}; programming through hierarchical tutoring, verification, classroom support, and retrieval grounding \citep{kargupta2024instruct,traver,kazemitabaar2024codeaid,kite,groher2026courseaware}; and structured science tutoring \citep{sharma2026convolearn}. Reliability also varies by language \citep{gupta2025reliablelanguages}. Whether methods calibrated on math survive domains with multiple valid answers or explanation-heavy outcomes is untested.

\subsection{Interaction Horizon}
\label{sec:data-gaps}

Most resources cover isolated problems or short dialogues, while personalization and learning are longitudinal claims. LongTutor uses extended histories for diagnosis and teaching \citep{li2026longtutor}, EduClaw-Bench evaluates repeated simulated interaction \citep{educlaw-bench}, and real-user work adds long deployments and delayed assessment \citep{oreopoulos2026khanmigo,oreopoulos2026mastery}. What is missing is not a longer conversation but repeated observation of the same learners linking history, tutor actions, delayed performance, and transfer. Joint annotation of learner state, action, safety, outcome, and simulator fidelity is rare, which is why end-to-end loop evaluation is so hard.

\section{Emerging Directions}
\label{sec:emerging}

Recent systems widen the loop past one text-only tutor and one learner. The validity questions do not change, but agents, modalities, languages, tools, and safety constraints each add another interface that can fail.

\subsection{Agentic and Multi-Agent Tutoring}
\label{sec:collaborative}

Agentic tutoring means one of two things. The first is functional decomposition, splitting assessment, planning, retrieval, teaching, reflection, and memory across components, as in GenMentor \citep{genmentor}, AgentTutor \citep{agenttutor}, DeepTutor \citep{zhao2026deeptutor}, and knowledge-regulated tutoring from slides \citep{zhangli2024slide2lecture}; modularity buys control but makes failure attribution harder. The second is genuine multi-party learning: MathVC \citep{yue2024mathvc}, SimClass \citep{zhang2025simclass}, and MAIC \citep{yu2026maic} add participants, so the policy must decide whom to address, when to intervene, and when to let peers talk. The two should be evaluated separately.

\subsection{Multimodal Tutoring}
\label{sec:multimodal}

Text-only tutors miss diagrams, handwriting, code traces, slides, and speech. Multimodal pedagogical instruction already appears in language learning \citep{multimodal-scaffolding}, and learner-centered architectures link dialogue to instructional videos \citep{li2025coggen}. The evaluation problem is decomposition: a bad response may come from a perceptual error, a wrong learner-state inference, or a poor pedagogical choice, and benchmarks should separate these rather than scoring the final turn.

\subsection{Retrieval-Grounded Tutors}
\label{sec:rag}

Retrieval is becoming part of the policy rather than a factuality patch. TutorLLM conditions retrieval on knowledge-tracing state \citep{tutorllm}; course-aware systems ground answers in instructional materials \citep{groher2026courseaware}; learner-state-aware RAG couples adaptation to retrieved evidence \citep{adenuga2026ragpedagogy}; and retrieval supports algorithm tracing \citep{kite}, lightweight tutoring interfaces \citep{hevia2025efficienttutor}, and just-in-time feedback \citep{lee2026jitfeedback}. Evaluation must separate retrieval quality, interpretation, and pedagogical use.

\subsection{Multilingual Tutoring}
\label{sec:multilingual}

Language tutoring already spans ESL \citep{kwon2024biped}, EFL writing \citep{han2024llmtutorEFL}, and proficiency-controlled dialogue \citep{almasi2025alignmentdrift}. The open question is whether the same policy holds across languages; cross-language educational evaluations show uneven reliability \citep{gupta2025reliablelanguages}. Multilingual work should test pedagogical invariance: whether diagnosis, scaffolding, answer withholding, and adaptation survive a change of interaction language.

\subsection{Safety and Alignment for Minors}
\label{sec:safety}

Tutoring means repeated interaction with learners, often children, so safety is part of the policy. Guardrails constrain inappropriate behavior and premature answer disclosure \citep{palchowdhury2024autotutor}, child-safety work shows age-sensitive risks need their own evaluation \citep{llm-safety-children}, training targets over-compliance and withholding \citep{yuan2026codae,pisan2026withhold}, and adversarial learners can still extract solutions \citep{zhao2026answerleakage}. Content safety and pedagogical safety differ: a response can be harmless as assistance yet harmful as teaching if it creates dependency or bypasses reasoning.

\end{document}